\pdfoutput=1

\documentclass[11pt]{article}

\usepackage[final]{acl}

\usepackage{times}
\usepackage{latexsym}
\usepackage{booktabs}
\usepackage{multirow}
\usepackage[T1]{fontenc}

\usepackage[utf8]{inputenc}

\usepackage{microtype}

\usepackage{inconsolata}

\usepackage{graphicx}
\usepackage{amsmath}
\usepackage{xspace}
\usepackage{colortbl}
\usepackage{tcolorbox}
\usepackage{xcolor}
\definecolor{Highlight}{rgb}{0.92,0.94,1}

\newcommand{\mname}{{\text G2}\xspace}
\newcommand{\aname}{{\text Guide-to-Generation}\xspace}
\newcommand{\pname}{{\text Guide}\xspace}
\newcommand{\posname}{{\text Diversity Guide}\xspace}
\newcommand{\negname}{{\text Dedupe Guide}\xspace}

\usepackage{algorithm}
\usepackage[noend]{algpseudocode}
\usepackage{multirow}
\usepackage{makecell}
\usepackage{subcaption}
\usepackage{amsmath}
\usepackage{amsfonts}
\usepackage{amssymb}
\usepackage{booktabs}

\title{G2: Guided Generation for Enhanced Output Diversity in LLMs}

\author{%
    Zhiwen Ruan$^{1}$, \
    Yixia Li$^{1}$, \
    Yefeng Liu$^{4}$,\
    Yun Chen$^{3}$,
    Weihua Luo$^{4}$\\
    \textbf{Peng Li}$^{2}$,
    \textbf{Yang Liu}$^{2}$,
    \textbf{Guanhua Chen}$^{1}\thanks{\ \ Corresponding author.}$ \\
    $^1$Southern University of Science and Technology, $^2$Tsinghua University  \\
    $^3$Shanghai University of Finance and Economics,
    $^4$Alibaba International Digital Commerce  \\
}

\begin{document}
\maketitle

\begin{abstract}

Large Language Models (LLMs) have demonstrated exceptional performance across diverse natural language processing tasks. However, these models exhibit a critical limitation in output diversity, often generating highly similar content across multiple attempts. This limitation significantly affects tasks requiring diverse outputs, from creative writing to reasoning. Existing solutions, like temperature scaling, enhance diversity by modifying probability distributions but compromise output quality.
We propose \textbf{Guide-to-Generation} (G2), a training-free plug-and-play method that enhances output diversity while preserving generation quality. G2 employs a base generator alongside dual Guides, which guide the generation process through decoding-based interventions to encourage more diverse outputs conditioned on the original query. Comprehensive experiments demonstrate that G2 effectively improves output diversity while maintaining an optimal balance between diversity and quality.


\end{abstract}

\section{Introduction}


Large Language Models (LLMs) have demonstrated remarkable capabilities in a wide range of natural language generation tasks~\cite{openai2024gpt4technicalreport,deepseekai2024deepseekv3technicalreport}. Despite their fluency and coherence, LLMs frequently generate generic, repetitive, or overly conservative outputs, even when explicitly instructed to produce random or diverse outputs \cite{zhang2024forcing,lanchantin2025diversepreferenceoptimization}. This limitation hinders their performance in tasks that demand diverse and informative responses, such as open-domain question answering, instruction following, and test-time scaling for reasoning~\cite{zhang2024edtimprovinglargelanguage}. 
Existing efforts to enhance generation diversity can be broadly categorized into training-time and decoding-time approaches. Training-based techniques modify the loss functions during supervised fine-tuning or reinforcement learning to encourage diverse outputs~\cite{zhang2024forcing,li2025preserving,lanchantin2025diversepreferenceoptimization}, but they often require additional training overhead and are customized for specific tasks, making them inefficient and inflexible.
Decoding-based approaches offer greater adaptability but face a fundamental \textbf{diversity–quality trade-off}. Sampling-based methods like temperature scaling increase variability by flattening the output distribution~\cite{peeperkorn2024temperaturecreativityparameterlarge, renze-2024-effect, zhu2024hot}, yet they fail to leverage previous responses and may only yield marginal diversity gains. Prompt-based methods enhance diversity by conditioning on prior generations, but repeated prompting often leads to semantic drift, undermining output quality~\cite{zhang2025noveltybenchevaluatinglanguagemodels}.

To address these limitations, we propose \aname (\mname), a training-free plug-and-play decoding strategy that enhances output diversity while preserving generation quality. \mname consists of three coordinated modules operating within the same LLM: a \textbf{base generator} focused on maintaining response quality, a \textbf{\posname} that encourages novel outputs, and a \textbf{\negname} that suppresses repetition. All modules share the same backbone and are differentiated by distinct prompting templates tailored to their specific roles.

To further promote diversity, we introduce a \textbf{center selection strategy (CSS)} that selects a small, semantically representative subset of prior generations to condition the guiding modules. Rather than using all previous outputs, which may contain redundant or overlapping content and thus dilute the guidance signal, CSS ensures that the diversity and deduplication prompts are anchored in distinct and non-redundant semantic cues. This targeted conditioning helps steer the generation away from previously explored regions of the output space, fostering greater novelty in subsequent responses.

Additionally, to avoid degradation in output quality due to overly aggressive guidance, \mname employs an \textbf{entropy-based selective intervention mechanism}. When the model exhibits high confidence in its token predictions, no intervention is applied; when uncertainty is high, guidance signals are selectively introduced. This mechanism allows \mname to intervene precisely where needed, enhancing diversity while preserving fluency and coherence.

Experimental results across creative and subjective generation, instruction-following, translation, summarization, and math tasks demonstrate that \mname significantly improves output diversity while maintaining high response quality, positioning it as a promising technique for enhancing the diversity of LLM-generated content.\footnote{Our code is publicly available at \url{https://github.com/sustech-nlp/emnlp25-g2}.}

\section{Related Work}

\subsection{Diverse Text Generation}


Diversity is essential in text generation for improving best-of-N performance~\cite{NEURIPS2020_1f89885d}, enabling synthetic dataset construction~\cite{raventos2023pretraining}, and generating alternatives for unsatisfactory initial outputs~\cite{zhang2024edtimprovinglargelanguage,garces-arias-etal-2024-adaptive}.

Existing methods for enhancing diversity fall into two categories: training-based and training-free. 
Empirical analyses have demonstrated that fine-tuning with standard cross-entropy loss tends to suppress generation diversity~\cite{o'mahony2024attributing,kim2025knowledge}. To mitigate this, recent works propose modifying the SFT objective to encourage more varied outputs~\cite{li2025preserving,zhang2024forcing}. Other work improves diversity by refining preference optimization objectives; for example, DivPO~\cite{lanchantin2025diversepreferenceoptimization} extends DPO to encourage more diverse generations.

Training-free approaches typically rely on decoding strategies such as temperature sampling~\cite{renze-2024-effect, zhu2024hot}, with some variants applying token-level temperature adjustments. For example, Entropy-Driven Temperature (EDT)\cite{zhang2024edtimprovinglargelanguage} dynamically adjusts temperature based on model entropy, while KLD-based methods\cite{chang2023kldivergenceguidedtemperaturesampling} use KL divergence between two models to guide temperature tuning. Although these techniques can increase diversity, they often ignore prior outputs and may degrade quality when overly high temperatures are applied.
Prompt-based methods attempt to enhance diversity by conditioning the model on its previous generations. While effective to some extent, repeated prompting can introduce semantic drift, leading to outputs that are less relevant or coherent relative to the original query~\cite{zhang2025noveltybenchevaluatinglanguagemodels}.

\subsection{Contrastive Decoding}

Contrastive Decoding (CD) enhances generation quality by leveraging likelihood differences between a strong and a weak language model~\cite{li-etal-2023-contrastive}. Recent CD methods fall into two main types: multi-model approaches~\cite{liu2024tuning,zhou2024weaktostrong,wu2024crossmodel,manevich-tsarfaty-2024-mitigating} and single-model techniques using prompt variation~\cite{CAD,preadd,sennrich-etal-2024-mitigating,leng2024mitigating}.
While multi-model setups offer stronger contrasts, single-model methods are more efficient and avoid vocabulary mismatches. For instance, Context-Aware Decoding~\cite{CAD} uses contrast between context-aware and context-free generations to reduce hallucinations, and Integrative Decoding~\cite{cheng2025integrative} uses prior outputs to enhance consistency. Our method introduces a novel use of contrastive signals between the original prompt and tailored prompts (diversifying and deduplication) to achieve both diversity and quality.

\section{Methodology}
\label{sec:methodology}

\begin{figure*}[t]
  \includegraphics[width=0.95\linewidth]{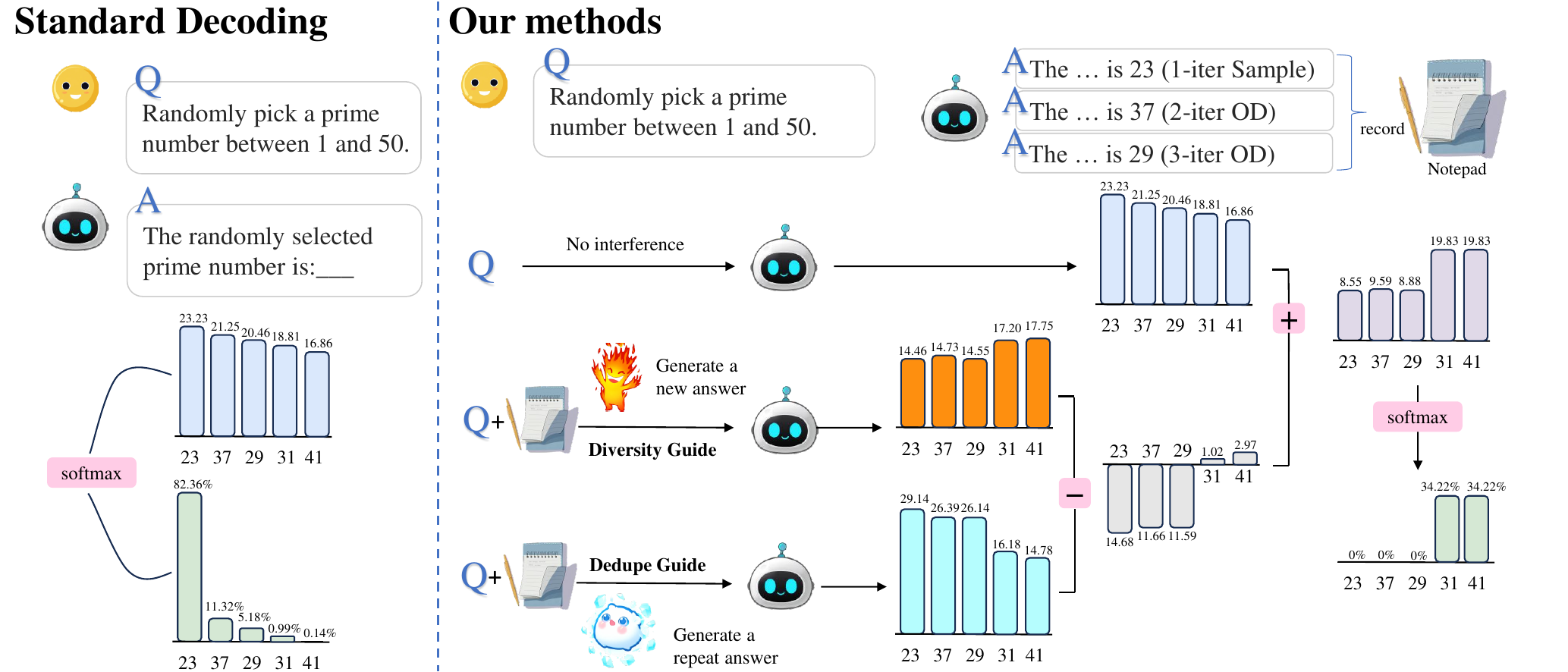}
  \caption{Comparison between standard decoding and our method \mname. Standard decoding often produces repetitive outputs, with certain tokens dominating due to peaked softmax distributions. \mname leverages Diversity Guide and Dedupe Guide to encourage diverse and novel generations. See Algorithm~\ref{alg:main_pseudocode} in the Appendix for details.}
  \label{fig:main}
  \vspace{-10pt}
\end{figure*}

Despite generating high-quality outputs, LLMs often struggle with response diversity, even when explicitly prompted for it~\cite{zhang2024forcing}.
To address this challenge, we propose \textbf{\mname}, a novel decoding strategy to dynamically guide the generation process towards greater diversity. The overall Algorithm~\ref{alg:main_pseudocode} is provided in the Appendix~\ref{app:sec:pseudocode_mname}.

\subsection{Overview}
Let $M$ denote the language model. For a given query $Q$ and prompt $\mathcal{P}$, our goal is to generate $N$ diverse answers $\{A_1, \dots, A_N\}$.
When generating the $i$-th answer, $A_i$, our objective is twofold: to ensure its accuracy in addressing the query $Q$ and its distinctiveness from the set of previously generated answers, $A_{<i} = \{A_1, A_2, \dots, A_{i-1}\}$. 

As shown in Figure \ref{fig:main}, the base generator, using model $M$, produces a logit vector $\mathbf{z}_t \in \mathbb{R}^{|V|}$ for the token at step $t$, where $|V|$ is the vocabulary size:
\begin{align}
 \mathbf{z}_t = M[\mathcal{P}, Q](\mathbf{x}_{<t})
\end{align}
The base generator is conditioned only on the initial prompt $\mathcal{P}$ and the query $Q$, and does not observe prior responses $A_{<i}$. This ensures that its generation remains focused on the query and unbiased by past outputs during this initial phase.

To steer the generation away from prior responses $A_{<i}$ and encourage novelty, we introduce two {\pname} components. Leveraging the strong instruction-following capabilities of LLMs, the base generator and these {\pname}s are implemented using the same language model $M$ but are provided with distinct instructional prompts and access to $A_{<i}$.
The Diversity {\pname} is prompted by $\mathcal{P}^+$ to generate content that diverges from $A_{<i}$, while the \textit{Dedupe \pname} is prompted by $\mathcal{P}^-$ to generate content that is similar to $A_{<i}$. Detailed prompt templates for $\mathcal{P}^+$ and $\mathcal{P}^-$ are provided in Appendix~\ref{sec:app_prompts}.
Their respective logits at token $t$ are computed as:
\begin{align}
\mathbf{z}_t^+ &= M[\mathcal{P}^+, Q, A_{<i}](\mathbf{x}_{<t}) \\
\mathbf{z}_t^- &= M[\mathcal{P}^-, Q, A_{<i}](\mathbf{x}_{<t})
\end{align}

To combine the guidance from both {\pname}s, we integrate their logits with the generator’s output to modulate the final token probability distribution:
\begin{align}
\mathbb{P}(X_t | \mathbf{x}_{<t}) = \text{softmax}(\mathbf{z}_t + \alpha_t(\mathbf{z}_t^+ - \mathbf{z}_t^-))
\label{equ:cal_next_token}
\end{align}
where $\mathbb{P}(X_t | \mathbf{x}_{<t})$ is the probability distribution for the next token $X_t$ given the preceding tokens $\mathbf{x}_{<t}$. 
The core directional guidance for diversification is provided by the contrastive signal $(\mathbf{z}_t^+ - \mathbf{z}_t^-)$. 
The Diversity {\pname} ($\mathbf{z}_t^+$) is prompted to assign higher logits to tokens that introduce novelty with respect to previous answers $A_{<i}$, and conversely, lower logits to tokens likely to cause repetition. The Dedupe {\pname} ($\mathbf{z}_t^-$), on the other hand, assigns higher logits to tokens that would closely mirror content in $A_{<i}$ (e.g., tokens `23', `37', `29' as illustrated in Figure~\ref{fig:main}). Therefore, the difference $(\mathbf{z}_t^+ - \mathbf{z}_t^-)$ serves to amplify the logits of novel tokens while suppressing those of repetitive ones, effectively guiding the model towards generating a distinct output.$\alpha_t$ is a dynamic scaling factor that controls the strength of this {\pname}'s influence on the final distribution. 


\subsection{Selective Intervention Strategy}
\label{subsec:selective_intervention}


While the {\pname} mechanism provides signals for diversity, determining the appropriate level and timing of intervention is crucial for maintaining quality. Overly aggressive or misapplied guidance can degrade output quality.
Therefore, \mname applies intervention selectively at each token generation step based on model uncertainty. We introduce an entropy-based gating strategy to compute the dynamic weighting factor $\alpha_t$ (from Equation~\ref{equ:cal_next_token}) for each token $X_t$.

We quantify model uncertainty using the entropy $H_t = H(\mathbb{P}(X_t | \mathbf{x}_{<t}))$ of its predictive distribution. Intervention is then gated by an entropy threshold $\beta$ and applied with a fixed strength $\theta$. If $H_t < \beta$, the model is considered confident, and no intervention occurs ($\alpha_t = 0$), preserving high-confidence predictions. If $H_t \geq \beta$, indicating sufficient uncertainty, the {\pname} mechanism is activated by setting $\alpha_t = \theta$. Formally, $\alpha_t$ is defined as:
$$ \alpha_t = \begin{cases} \theta & \text{if } H(\mathbb{P}(X_t | \mathbf{x}_{<t})) \geq \beta \\ 0 & \text{if } H(\mathbb{P}(X_t | \mathbf{x}_{<t})) < \beta \end{cases} $$
This strategy offers advantages over raw entropy or fixed-weight interventions by better preserving quality. Specifically, it avoids intervening on confident (low-entropy) predictions, and the cap $\theta$ prevents degradation from excessive interventions, particularly on high-entropy tokens. The ablation study (Section~\ref{sec:ablation}, Figure~\ref{fig:ablation}) confirms that these alternative strategies result in quality drops.



The threshold $\beta$ (fixed at $0.1$ in all our experiments) and intervention strength $\theta$ (e.g., 0.3 or 0.5, selected based on validation experiments detailed in Appendix~\ref{sec:app_hyperparam_search}; $\theta=0$ disables all intervention) help manage the diversity-quality trade-off. This selective application strategically promotes diversity primarily at points of model uncertainty, thereby balancing novelty with coherence and quality.

\subsection{Representative Prior Response Sampling}
\label{subsec:representative_sampling}


When generating the $n$-th response, $A_n$, conditioning the {\pname}s on all $n-1$ prior responses ($A_1, A_2, \dots, A_{n-1}$) can result in excessively long prompts. Such lengthy contexts may degrade model performance and are often unnecessarily verbose due to semantic redundancies frequently present within the set of prior responses, $A_{<n}$. Thus, selecting a diverse and representative subset of these prior responses is a more effective strategy.

To facilitate this selection, we first derive semantic embeddings for each prior response $A_j \in A_{<n}$. To avoid reliance on external models, we utilize the base LLM itself for this task. Specifically, each response $A_j$ is processed using the instructional prompt: "This sentence: \{$A_j$\} means in one word:", and its sentence embedding is extracted from a designated hidden state of the LLM's output. 
The confluence of a typically small set of prior responses and their high-dimensional embeddings often renders conventional clustering methods suboptimal for selecting a genuinely diverse and representative subset of exemplars.

We therefore employ a greedy iterative method, termed the Center Selection Algorithm, to curate this representative subset. The algorithm initializes by selecting an initial response from $A_{<n}$ (e.g., the most recent or a random one). Subsequently, it iteratively adds the response from the remaining unselected pool that exhibits the maximum dissimilarity (e.g., maximizing the minimum cosine dissimilarity based on their embeddings) to any response already included in the representative set. This process continues until a predefined number of representative responses are selected. This condensed subset then serves as the context of prior outputs for the {\pname} prompts.

\section{Experiments}

\begin{table*}[!htb]
    \centering
    \resizebox{0.85\textwidth}{!}{
    \begin{tabular}{l|ccc|ccc}
        \toprule
        \textbf{Methods} & \textbf{Div-BLEU} &  \textbf{EAD} &  \textbf{Sent-Bert} &  \textbf{Diversity} ($\uparrow$) &\textbf{Distinct} ($\uparrow$) & \textbf{Quality} ($\uparrow$)  \\
        \midrule
        Claude-3.5 Sonnet & 36.27 & 47.96 & 19.68 &	30.90  & 1.50 & 8.60  \\
        gpt-4o &36.64 & 50.08 & 21.53 & 32.45  & 1.95 & 9.01 \\
        Llama-3.3-70B-Instruct & 32.24 & 44.66 & 17.78 & 28.12 & 1.70 & 8.57 \\
       \bottomrule \toprule
        Llama3-8B-Instruct  & 53.59 & 69.22 & 29.4 & 45.40  & 4.02 & 7.97 \\ \midrule
        \textit{w.} Temperature (T=1.3) & 62.29 & 73.61 & 34.44 & 51.20  (\textcolor{olive}{+5.80})  & 4.63 (\textcolor{olive}{+0.61}) & \underline{7.69} (\textcolor{olive}{-0.28})  \\
        \textit{w.} Top-P (T=1.5, P=0.95) & 64.60 & 75.47 & 36.44 & \underline{53.24} (\textcolor{olive}{+7.84}) & 5.17 (\textcolor{olive}{+1.15}) & 7.61 (\textcolor{olive}{-0.36}) \\
        \textit{w.} Top-K (T=1.5, K=10) & \underline{64.65}	& \underline{75.86}	& 36.17 & 53.21 (\textcolor{olive}{+7.81}) &	5.08 (\textcolor{olive}{+1.06}) & 7.63 (\textcolor{olive}{-0.34})  \\
        \textit{w.} Min-P (T=1.5, P=0.01) & 65.24 & 75.09 &	35.62 & 52.89 (\textcolor{olive}{+7.49}) & 5.25 (\textcolor{olive}{+1.23}) & 7.61 (\textcolor{olive}{-0.36}) \\
        \textit{w.} EDA (T=1.5, $\theta$=0.1) & 60.81 & 73.93 & 33.73	& 50.55 (\textcolor{olive}{+5.15}) & 5.06 (\textcolor{olive}{+1.04}) & 7.61 (\textcolor{olive}{-0.36})  \\
        \textit{w.} Diverse Prompt (T=1.0) & 57.40 & 73.90 & \textbf{40.53} & 53.09 (\textcolor{olive}{+7.69}) &	\textbf{6.58} (\textcolor{olive}{+2.56}) & 4.73 (\textcolor{olive}{-3.24})  \\
        \midrule

        \rowcolor{Highlight}
        \textit{w.} \mname ($\theta$=0.3) & \textbf{64.72} &	\textbf{78.01} &	\underline{37.91} &	\textbf{54.64} (\textcolor{olive}{+9.24}) & 	\underline{5.80} (\textcolor{olive}{+1.78}) &	\textbf{7.79} (\textcolor{olive}{-0.18}) \\

        \bottomrule \toprule
        Qwen2.5-7B-Instruct & 56.78 & 69.42 & 32.84 & 47.97	& 3.60 & 7.50 \\ \midrule
        \textit{w.} Temperature (T=1.3) & \textbf{66.25} & 75.04 & 34.73 & 52.69 (\textcolor{olive}{+4.72}) &	4.60 (\textcolor{olive}{+1.00}) & 7.03 (\textcolor{olive}{-0.47})  \\
        \textit{w.} Top-P (T=1.5, P=0.95) & 64.76 &	75.72 &	36.20 &	\underline{53.22} (\textcolor{olive}{+5.25}) & 	4.84 (\textcolor{olive}{+1.24}) &	6.99 (\textcolor{olive}{-0.51})  \\
        \textit{w.} Top-K (T=1.5, K=10) & 64.51 &	\underline{76.80} &	35.30 &	52.98 (\textcolor{olive}{+5.01}) & 	4.90 (\textcolor{olive}{+1.30}) &	7.20 (\textcolor{olive}{-0.30})  \\
        \textit{w.} Min-P (T=1.5, P=0.01) & 66.29 &	75.0 &	34.54 & 52.60  (\textcolor{olive}{+4.63}) & 4.82 (\textcolor{olive}{+1.22}) & 6.99 (\textcolor{olive}{-0.51}) \\
        \textit{w.} EDA (T=1.3, $\theta$=0.1) & 62.82 &	73.28 &	34.49 &	51.27 (\textcolor{olive}{+3.30}) & 	4.50 (\textcolor{olive}{+0.90}) &	\textbf{7.21} (\textcolor{olive}{-0.29}) \\
        \textit{w.} Diverse Prompt (T=1.0) & 56.38 &	74.12 &	\underline{37.78} &	51.52 (\textcolor{olive}{+3.55})  & 	\textbf{6.29} (\textcolor{olive}{+2.69}) &	4.03 (\textcolor{olive}{-3.47})   \\
        \midrule
        \rowcolor{Highlight}
        \textit{w.} \mname ($\theta$=0.15) & \underline{65.53} & \textbf{78.79} & \textbf{38.04} & \textbf{55.10} (\textcolor{olive}{+7.13})	& \underline{5.46} (\textcolor{olive}{+1.86}) & \textbf{7.21} (\textcolor{olive}{-0.29}) \\
        \bottomrule
    \end{tabular}
    }
    \caption{Performance comparison of decoding methods on the NoveltyBench benchmark. While the Diverse Prompt baseline achieves higher Distinct diversity scores, its Quality significantly deteriorates compared to our method (\mname). For an expanded comparison across more parameter settings, see Figure \ref{fig:app_novelty} in Appendix~\ref{sec:app_novelty_bench}.}
    \label{tab:novelty-bench}
    \vspace{-10pt}
\end{table*}


This section presents comprehensive experiments evaluating \mname across diverse NLP tasks (creative generation, instruction-following, translation, and summarization) for a multifaceted assessment.

\subsection{Experimental Setup}
\label{sec:exp_setup}

This section outlines the evaluation protocols, including diversity metrics, baseline methods, and implementation details employed across our experiments. Task-specific quality metrics are detailed within their respective benchmark discussions.

\paragraph{Diversity Evaluation}
We assess generation diversity using a suite of established metrics, ensuring a comprehensive evaluation from both lexical and semantic perspectives. Higher values consistently indicate greater diversity.

\noindent $\bullet$ \textbf{Div-BLEU}: Calculated as $1 - \text{Self-BLEU}$ \cite{zhu2018texygen}.

\noindent $\bullet$ \textbf{EAD (Expectation-Adjusted Distinct N-grams)}: Counts distinct 1- to 5-grams, adjusted to mitigate bias from shorter outputs \cite{li-etal-2016-diversity,liu-etal-2022-rethinking}.

\noindent $\bullet$ \textbf{Sent-BERT}: Measures semantic diversity as $1 - \text{average cosine similarity}$ between Sentence-BERT embeddings of the responses \cite{kirk2024understanding}.

These three metrics are aggregated into a composite diversity score, diversity (Div), which gives equal (50\%) weight to statistical diversity (EAD and Div-BLEU combined) and semantic diversity (Sent-BERT). The formulation is:
\begin{align}
Div = \frac{\text{EAD} + \text{Div-BLEU}}{4} + \frac{\text{Sent-BERT}}{2}
\end{align}

\paragraph{Baselines}
We compare our approach against the following baselines. All methods, including ours, utilize Llama-3-8B-Instruct \cite{llama3modelcard} as the backbone LLM to ensure fair comparison.

\noindent $\bullet$ \textbf{Fixed Temperature Sampling}:
Standard sampling with a fixed temperature T, Top-K (50), and Top-P (1.0). $T$ is varied for trade-off analysis, where $\text{T} \in \{1.0, 1,1, 1.2, 1.3, 1,4, 1.5\}$.

\noindent $\bullet$ \textbf{Top-P Sampling}:
Employs a high temperature (T=1.5) to encourage diversity, with the Top-P ($\text{P} \in \{0.8,0.85, 0.9, 0.95\}$) threshold varied to modulate quality.

\noindent $\bullet$ \textbf{Top-K Sampling}:
Similar to Top-P, it uses a high temperature (T=1.5), varying the Top-K ($\text{K} \in \{5, 10, 20, 40\}$) parameter.

\noindent $\bullet$ \textbf{Min-P Sampling}:
A dynamic truncation method that adjusts the sampling threshold based on the model’s confidence \cite{minh2025turning}, which uses a high temperature (T=1.5), varying the $\text{Min-P} \in \{0.01,0.03,0.05,0.07\}$.

\noindent $\bullet$ \textbf{Diverse Prompt}:
Leverages a prompting strategy \cite{zhang2025noveltybenchevaluatinglanguagemodels} that conditions the LLM on previous outputs to encourage varied responses.

\noindent $\bullet$ \textbf{EDT}:
This method \cite{zhang2024edtimprovinglargelanguage} dynamically adjusts temperature based on token-level entropy. It uses an initial temperature $T$ and a fixed adjustment strength $\theta = 0.1$ (consistent with the original work). For WMT‘14 and XLSum, we set $T \in \{1.2, 1.4, 1.6, 1.8, 2.0\}$ to allow broader exploration; for other tasks, $T \in \{1.0, 1.3, 1.5\}$.

\paragraph{Details}
Unless specified otherwise, all methods use Top-K=50 and Top-P=1.0. We generate ten outputs per query for NoveltyBench (following its standard protocol) and five per instance for other benchmarks. For all methods, the first of the multiple outputs is obtained via greedy decoding to establish a consistent quality baseline, while subsequent outputs are generated using the respective method's sampling strategy. 
For \mname, the intervention strength parameter $\theta$ is selected from the set $\{0.15, 0.3, 0.5, 0.7\}$, based on validation experiments detailed in Appendix~\ref{sec:app_hyperparam_search}, and the temperature is fixed at 1.0. All experiments are conducted on NVIDIA A100-80G GPU.

\subsection{Creative and Subjective Generation Task}
\label{sec:novelty_task}

\subsubsection{Experimental Setup}
\paragraph{Benchmark}
We employ NoveltyBench Curated \cite{zhang2025noveltybenchevaluatinglanguagemodels}, a benchmark specifically designed with prompts that elicit multiple valid and diverse answers. Its curated dataset spans creative writing, randomness, factual knowledge, and subjective opinion generation, making it ideal for assessing diversity in open-ended tasks.

\paragraph{Diverisity Metric}
In addition to the common metrics defined in Section~\ref{sec:exp_setup}, we report NoveltyBench's specialized \textbf{Distinct} metric, which quantifies the number of unique equivalence classes among $N$ generated responses.

\paragraph{Quality Metric}
Following NoveltyBench, generation quality is assessed using Skywork-Reward-Gemma-2-27B-v0.2 \cite{liu2024skywork} as an automated reward model. We report the average reward score over $N$ responses for each query.

\subsubsection{Main Results}
\label{sec:noveltybench_results}

Table~\ref{tab:novelty-bench} summarizes the performance of various models and decoding strategies on NoveltyBench. We highlight several key observations:

State-of-the-art proprietary models (Claude-3.5 Sonnet, GPT-4o) and large open-source models (Llama-3.3-70B-Instruct) achieve high generation quality. However, their output diversity is notably constrained, evidenced by low Distinct scores (1.50, 1.95, 1.70). This underscores the prevalent challenge of diversity in highly capable models, corroborating observations by \citet{lanchantin2025diversepreferenceoptimization}.

Conventional diversity-enhancement techniques, such as increasing sampling temperature (e.g., T=1.3), can improve diversity and distinct score by 5.8 and 0.61 points, respectively. However, this typically incurs a quality penalty, exemplified by a 0.28 point decrease in the quality score. While Top-K/P and Min-P sampling aim to mitigate this quality degradation at higher temperatures, they often achieve this by tempering the diversity gains. EDT~\cite{zhang2024edtimprovinglargelanguage} provides fine-grained, token-level temperature adjustments; yet it fails to leverage previous responses and may only yield marginal diversity gains. These results highlight the ongoing challenge of simultaneously maximizing diversity and quality.

The Diverse Prompt strategy, by conditioning on previous outputs, substantially boosts diversity metrics. However, this comes at a steep cost to output quality (e.g., quality score plummets 3.47 points), rendering it impractical for scenarios demanding high quality. This highlights the difficulty of maintaining relevance and coherence when aggressively prompting for novelty over multiple iterations.

In contrast, \textbf{\mname} lies near the Pareto frontier, striking an effective trade-off between diversity and quality. On Llama3-8B-Instruct, \mname reaches the highest diversity score and a strong distinct score, while maintaining high quality. Similar patterns are observed on Qwen2.5-7B-Instruct~\cite{qwen2.5}, where \mname consitently outperforms other methods in balancing novelty with coherence. As shown in Figure~\ref{fig:app_novelty} (Appendix~\ref{sec:app_novelty_bench}), \mname reliably occupies the upper-right region of the diversity-quality space, demonstrating its effectiveness in navigating the trade-off between generating varied outputs and preserving fluency and relevance.

\subsection{Instruction-Following Task}
\label{sec:if_task}

\begin{table}[!htb]
    \centering
    
    \resizebox{\columnwidth}{!}{
    \begin{tabular}{l|ccc|cc}
        \toprule
        \textbf{AlpcaEval 2.0} & \textbf{Div-BLEU} &  \textbf{EAD} &  \textbf{Sent-Bert} &  \textbf{Diversity} ($\uparrow$) & \textbf{LCWR} ($\uparrow$)  \\
        \midrule
        Llama3-8B-Instruct  & 53.52 & 70.13 & 18.46 & 40.14 &
        32.01 \\
        \midrule
        \textit{w.} Temperature (T=1.3) & 62.02 & 75.54 &	20.75 & 44.77 & \underline{28.81}   \\
        \textit{w.} Top-P (T=1.5, P=0.95) & 62.55 &	75.87 &	20.98 & 45.10  & 27.96   \\
        \textit{w.} Top-K (T=1.5, K=10) & 63.71 &	76.64 & 	20.85 & 45.51  & 27.92  \\
        \textit{w.} Min-P (T=1.5, P=0.01) & 64.20 & 	76.73 & 20.84 & 45.65   & 27.63   \\
        \textit{w.} EDA (T=1.5, $\theta$=0.1) & 65.04 &	77.07 &	21.43& \underline{46.24} &25.64   \\

        \rowcolor{Highlight}
        \textit{w.} \mname ($\theta$=0.5) & 64.10 &	75.91 &	24.00 & \textbf{47.00}  &  \textbf{29.20}  \\
        \bottomrule \toprule

        \textbf{MT-Bench} & \textbf{Div-BLEU} &  \textbf{EAD} &  \textbf{Sent-Bert} &  \textbf{Diversity} ($\uparrow$) & \textbf{Score} ($\uparrow$) \\ \midrule
        Llama3-8B-Instruct &55.47 & 70.36 &	22.65 & 42.78 & 7.18  \\
        \midrule
        \textit{w.} Temperature (T=1.3) & 62.93 & 74.73 & 25.38 & 47.11 & \textbf{6.96}  \\
        \textit{w.} Top-P (T=1.5, P=0.95) & 64.76 &	76.19 &	25.86 & 48.17 & 6.84   \\
        \textit{w.} Top-K (T=1.5, K=10) & 64.21 &	76.13 &	25.88 & 48.03 & 6.89   \\
        \textit{w.} Min-P (T=1.5, P=0.01) & 66.07 &	77.21 &	26.16 & 48.90    & 6.86    \\
        \textit{w.} EDA (T=1.5, $\theta$=0.1) & 66.19 &	77.60 &	26.08 & 48.99  & 6.75  \\
        \rowcolor{Highlight}
        \textit{w.} \mname ($\theta$=0.5) & 66.55	 & 76.06 &	27.60 & \textbf{49.45} &  \underline{6.92}  \\
        \bottomrule
    \end{tabular}
    }
    \caption{Performance comparison on instruction-following tasks. For an expanded comparison across more parameter settings, see Figure \ref{fig:app_alpaca_mt_bench} in Appendix~\ref{sec:app_if_task}}
    \label{tab:alpaca_mtbench}
    \vspace{-15pt}
\end{table}

\begin{figure*}[htbp]
\centering
\begin{subfigure}{0.49\linewidth}
    \centering
    \includegraphics[width=0.95\linewidth]{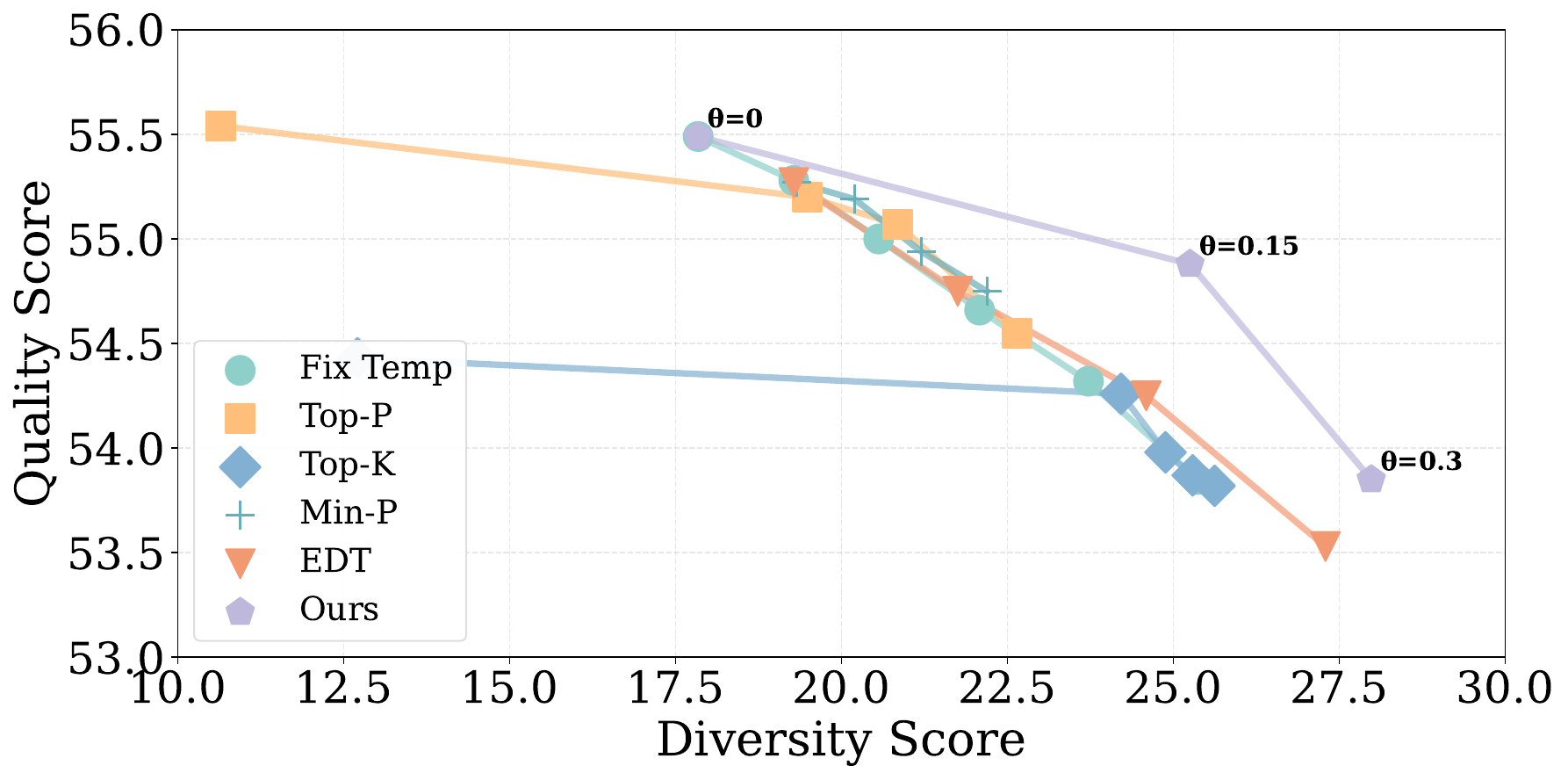}
    \caption{WMT'14 German-English}
    \label{1th}
\end{subfigure}
\centering
\begin{subfigure}{0.49\linewidth}
    \centering
    \includegraphics[width=0.95\linewidth]{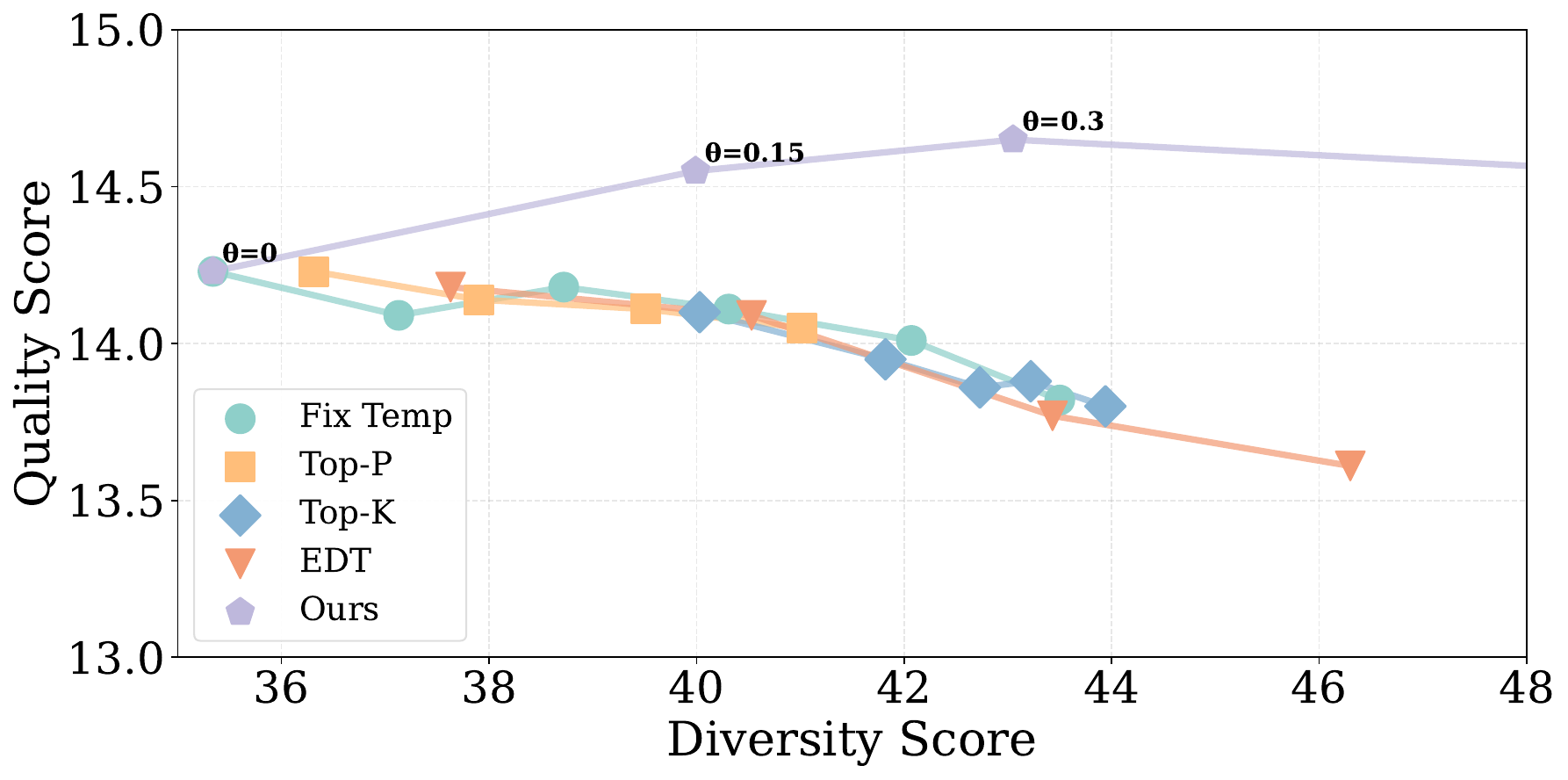}
    \caption{XLSum}
    \label{12th}
\end{subfigure}
\caption{Diversity-quality curves on WMT’14 and XLSum between \mname and other baseline under different settings.}
\label{fig:trans_summay}
\end{figure*}

\subsubsection{Experimental Setup}


\paragraph{Benchmark}
We evaluate our methods on two widely used instruction-following benchmarks: AlpacaEval 2.0 \cite{dubois2024lengthcontrolled} and MT-Bench \cite{zheng2023judging}. AlpacaEval 2.0 is a single-turn dialogue task where the evaluation metric is the length-controlled win rate (LCWR), which adjusts for the inherent length bias of the judging model. MT-Bench is a two-turn dialogue task, where the evaluation metric is the average score.

\paragraph{Quality Metrics}
We use GPT-4o-2024-08-06 \cite{openai2024gpt4technicalreport} as the judge model and calculate the average score of $N$ responses per query.

\subsubsection{Main Results}

Table~\ref{tab:alpaca_mtbench} summarizes results for instruction-following tasks, with further visualizations in Figure~\ref{fig:app_alpaca_mt_bench} (Appendix \ref{sec:app_if_task}). These findings reinforce \mname's consistent advantages.

On AlpacaEval 2.0, \mname achieves the highest diversity among the tested methods while maintaining a strong quality score (LCWR). This demonstrates a more effective diversity-quality balance compared to baselines like Top-K sampling or Temperature scaling, which exhibit a more pronounced decline in quality when attempting to reach similar high levels of diversity. Similarly, on MT-Bench, \mname again leads in Diversity while maintaining a competitive quality score. Compared to Temperature scaling (T=1.3), \mname's score is only 0.04 lower, yet its diversity is 2.34 points higher. Moreover, both the quality and diversity of \mname are higher than other approaches, such as EDA.

These results are corroborated by visual analyses (Figure~\ref{fig:app_alpaca_mt_bench}), where \mname's performance curve consistently occupies the desirable top-right region of diversity-quality plots, signifying its superior trade-off characteristics relative to existing baselines.

\subsection{Translation and Summarization Task}
\label{sec:trans_summary}

\subsubsection{Experiment Setup}
\label{sec:trans_exp_setup}

\paragraph{Benchmark}
We evaluate translation on WMT'14 German-English (de$\rightarrow$en; 3,003 sentence pairs) \cite{bojar-EtAl:2014:W14-33} and summarization on 1,000 English instances from XLSum \cite{hasan-etal-2021-xl}.

\paragraph{Metrics}
Translation quality is assessed by the average of BLEU\footnote{\url{https://www.nltk.org}} \cite{papineni-etal-2002-bleu} and COMET\footnote{\url{https://github.com/Unbabel/COMET}} \cite{rei-etal-2020-comet}. Summarization quality uses the average of ROUGE-1, ROUGE-2, and ROUGE-L\footnote{\url{https://pypi.org/project/rouge-score}} \cite{lin2004rouge}. Diversity is evaluated as in Section~\ref{sec:exp_setup}.

\subsubsection{Main Results}



Figure~\ref{fig:trans_summay} illustrates the diversity-quality relationship for \mname and baselines on WMT'14 de$\rightarrow$en and XLSum. Across both benchmarks, \mname consistently operates in the top-right region, signifying a superior trade-off between generation quality and output diversity compared to other evaluated methods.

It is noteworthy that translation and summarization tasks are more constrained than the open-ended tasks like creative generation or instruction-following. 
These tasks necessitate strict adherence to source content, rendering them less sensitive to standard diversity-enhancement techniques like temperature scaling at moderate settings. Consequently, standard techniques like temperature scaling often need to be set to considerably higher values (e.g., T = 1.5) to yield substantial diversity gains in these contexts. However, such aggressive temperature scaling typically leads to a severe degradation in output quality. This context underscores the effectiveness of \mname in successfully navigating this challenging landscape to improve diversity while preserving the fidelity and quality of the generated translations and summaries.

\section{Analyses}

\subsection{Ablation Study}
\label{sec:ablation}

\begin{figure}[t]
  \includegraphics[width=\linewidth]{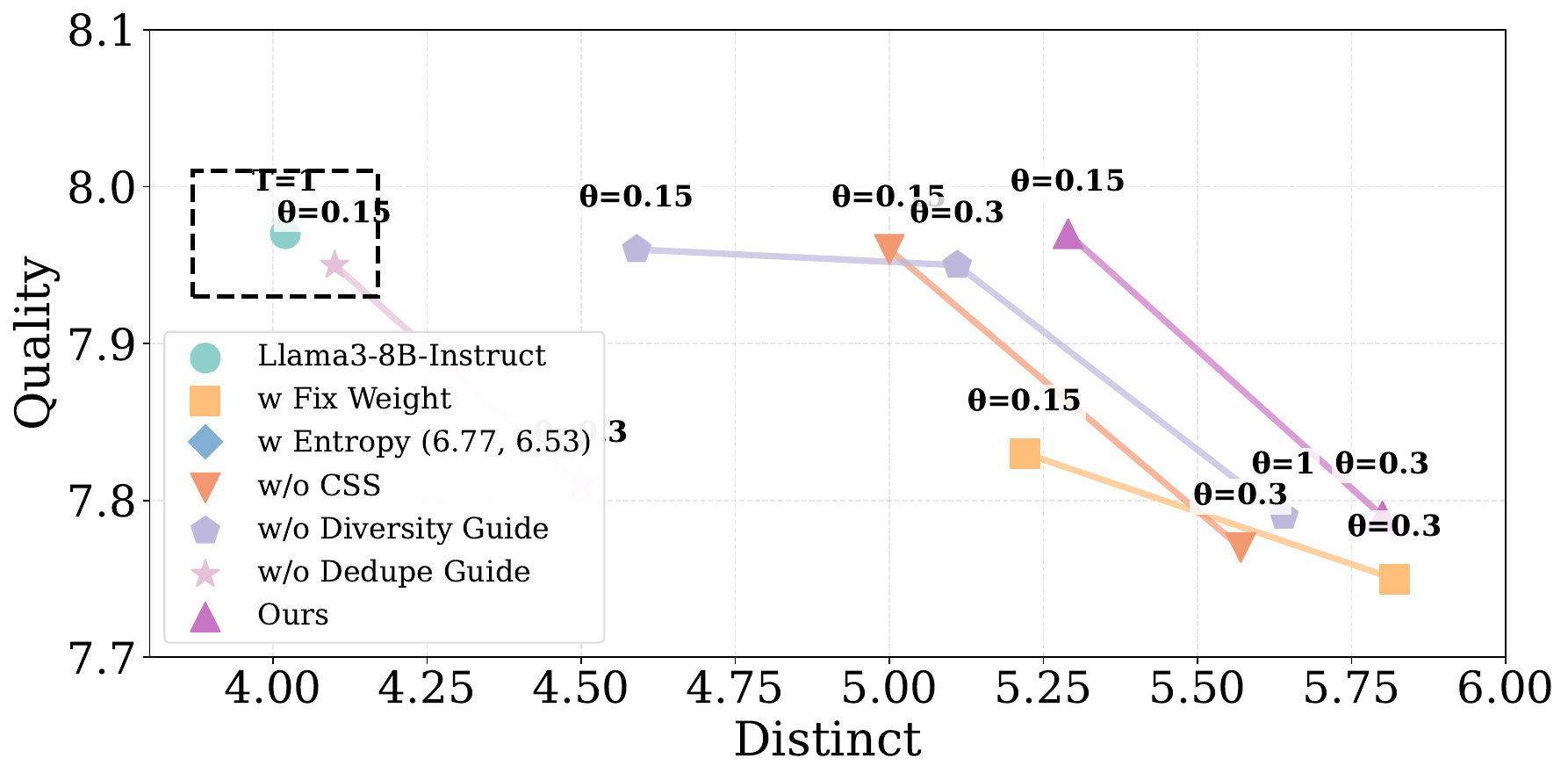}
  \caption{Ablation results of intervention strategy, center selection, and guide components on NoveltyBench.}
  \label{fig:ablation}
\end{figure}

We conduct ablation studies to evaluate contributions of key components in \mname on NoveltyBench.

\paragraph{Intervention Strategy}
We compare three strategies: \textit{w Fix Weight}, which intervenes on all tokens with fixed weights ($\theta=0.15$ and $0.3$); \textit{w Entropy}, which uses raw entropy as weights; and \mname, which selectively intervenes based on entropy with capped strength. As shown in Figure~\ref{fig:ablation}, \textit{w Fix Weight} achieves similar diversity to \mname under the same $\theta$, but \mname yields better quality, especially at $\theta=0.15$, while intervening on approximately 45\% fewer tokens. \textit{w Entropy} improves diversity but significantly reduces quality due to overly large weights on high-entropy tokens, demonstrating the advantage of selective intervention in \mname.

\paragraph{Center Selection Algorithm}
We compare \mname with and without the Center Selection Strategy (CSS), where \textit{w/o CSS} randomly selects past responses instead of using representative ones. As shown in Figure~\ref{fig:ablation}, \mname achieves higher diversity under similar quality, indicating that selecting representative responses via CSS helps guide the model to generate more novel content.

\paragraph{Guide}
We ablate the two \pname modules: the \posname and the \negname. \textit{w/o \posname} removes the diversity constraint, while \textit{w/o \negname} removes the dedupe constraint. As shown in Figure~\ref{fig:ablation}, both variants reduce diversity compared to \mname, with a more notable drop when removing \negname. We hypothesize that this is because prompting the model to generate repeated tokens (via consistency prompts) is easier than generating novel ones, making it more effective for contrastive decoding to suppress those repetitions and thereby enhance diversity. Overall, using both Guides leads to the best trade-off.

\subsection{Enhancing Mathematical Reasoning via Diverse Candidate Generation}

\begin{figure}[t]
  \includegraphics[width=\linewidth]{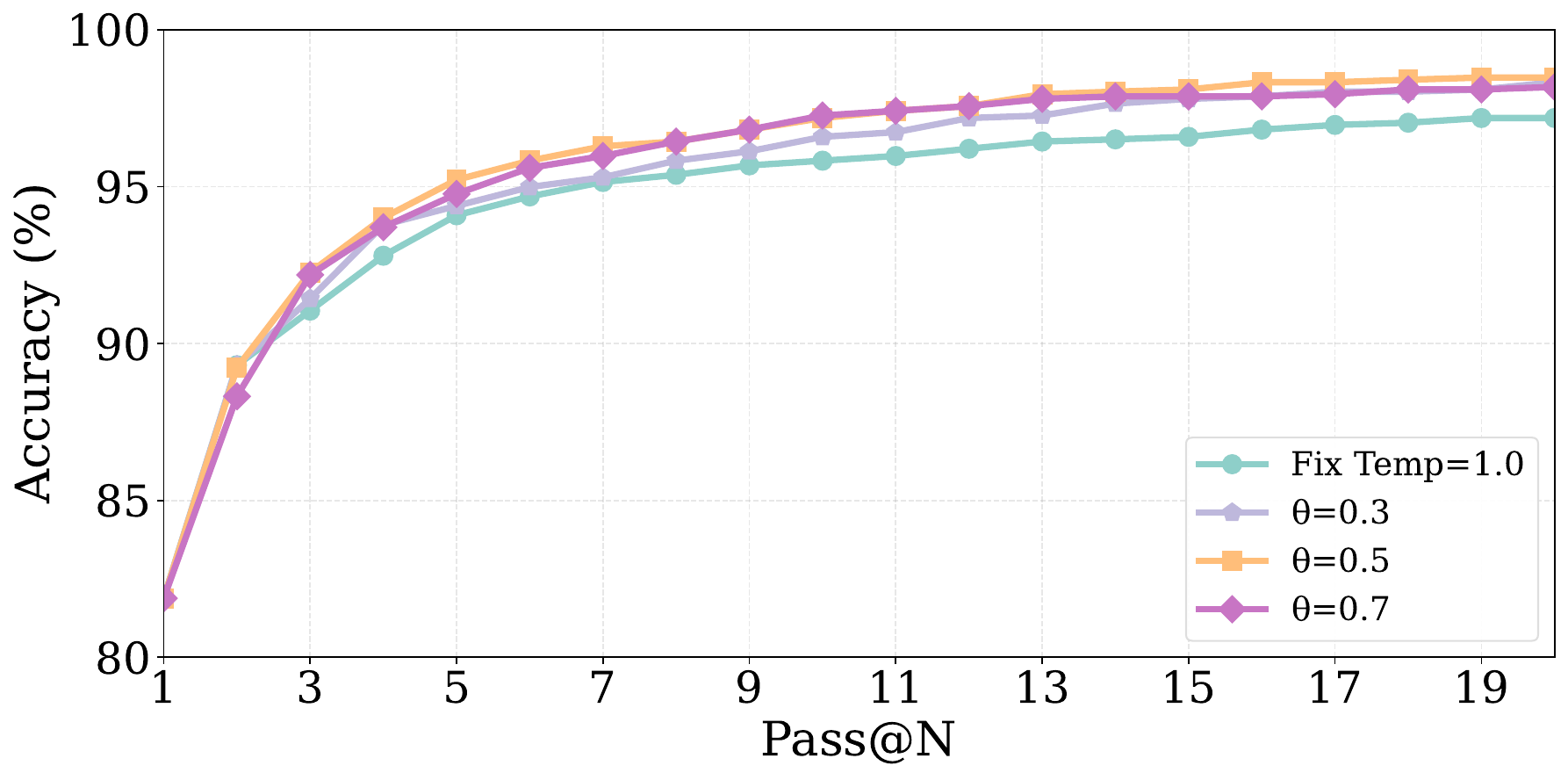}
  \caption{Comparison of Pass@N accuracy on GSM8K between our method and the baseline.}
  \label{fig:passn}
  \vspace{-15pt}
\end{figure}

For complex reasoning tasks such as mathematical problem-solving, relying on a single generation path may not consistently yield the correct solution. Generating a diverse set of candidate solutions effectively utilizes additional inference-time computation, a technique well-documented to enhance LLM output quality, particularly for reasoning tasks \cite{welleck2024from,snell2025scaling}.
We evaluate mathematical reasoning capabilities using the GSM8K benchmark \cite{cobbe2021gsm8k} with Llama-3-8B-Instruct. For each query, we sample \(N\) outputs and report Pass@N accuracy, which measures whether at least one correct answer appears among the samples. Unlike the BoN approach with a reward model, Pass@N reduces the reliance on the reward model's performance, providing a more direct assessment of the model’s ability to generate correct answers.

The hyperparameter $\theta$ allows for a nuanced control: \(\theta = 0.3\) setting tends to prioritize the quality of individual responses, whereas \(\theta = 0.7\) steers the model towards greater diversity among responses. As shown in Figure~\ref{fig:passn}, our method consistently outperforms the baseline when \(N \geq 3\), demonstrating its effectiveness in improving reasoning performance.
The high Pass@N performance illustrated in Figure~\ref{fig:passn} underscores that a strategic balance between the intrinsic quality of individual solutions and the collective diversity across multiple candidates is paramount for maximizing performance on such reasoning tasks.

\subsection{Correlation Between Generation Probability and Diversity}
\begin{table}[]
\centering
\resizebox{0.95\columnwidth}{!}{
\begin{tabular}{lcccccc}
\toprule
\textbf{Methods} & \textbf{Temp} & \textbf{$\theta$} & \textbf{ATLP} & \textbf{Quality} & \textbf{Diversity} \\
\midrule
Fix Temp & 1.0 & - & -0.64 & 55.49 & 17.84 \\
\mname & 1.0 & 0.15 & -0.72 & 54.88 & 25.25 \\
Fix Temp & 1.3 & - & -0.74 & 54.66 & 22.08 \\
\mname & 1.0 & 0.3 & -0.83 & 53.85 & 27.98 \\
Fix Temp & 1.5 & - & -0.86 & 53.85 & 25.36 \\
\bottomrule
\end{tabular}
}
\caption{
    Comparison of average token log-probability (ATLP), output quality, and diversity on WMT'14.
}
\label{tab:prob}
\vspace{-15pt}
\end{table}
To better understand the underlying reason why \mname improves generation diversity, we examine whether it achieves this by deviating from the model’s output distribution or by exploring diverse yet likely generation paths that remain aligned with the distribution. We conduct experiments on the WMT'14 German-English translation task (following the setup in Section~\ref{sec:trans_summary}) and compare \mname with fixed-temperature sampling.

We use the average token log-probability (ATLP) to measure how closely generated outputs align with the model's predicted distribution. As shown in Table~\ref{tab:prob}, ATLP is positively correlated with output quality across decoding strategies—higher-likelihood generations generally exhibit better quality. Within the same decoding method, a common trade-off is observed: increasing diversity (e.g., through higher temperature) typically leads to lower ATLP and reduced quality. However, this pattern does not always hold across different methods. Notably, \mname with $\theta = 0.15$ yields both higher ATLP and greater diversity than fixed-temperature decoding with $T = 1.3$.
These findings indicate that \mname promotes diversity by following varied yet probable generation paths, rather than deviating from the model’s distribution.

\subsection{Efficiency Analysis}

\begin{table}[]
\centering
\resizebox{0.95\columnwidth}{!}{
\begin{tabular}{lcc}
\toprule
\textbf{Method} & \textbf{Prior Responses (N)} & \textbf{Latency ($\times$ baseline)} \\
\midrule
Standard temperature sampling & -- & 1.00 \\
Top-P / Top-K / Min-P         & -- & $\approx$ 1.00 \\
EDT                           & -- & $\approx$ 1.05 \\
\midrule
G2 (sequential) & 1 / 2 / 3 & 2.52 / 2.66 / 2.67 \\
G2 (parallel)   & 1 / 2 / 3 & 1.19 / 1.26 / 1.31 \\
\bottomrule
\end{tabular}
}
\caption{
Relative inference latency of G2 compared to standard decoding methods on NoveltyBench.
}
\label{tab:efficiency}
\end{table}

To better understand the runtime characteristics of G2, we benchmarked its \textbf{inference latency} relative to common decoding strategies on NoveltyBench. Since G2 requires the base generator, the Diversity Guide, and the Dedupe Guide, the additional computation cost is a key factor to evaluate. Table~\ref{tab:efficiency} summarizes the latency results, reported relative to standard temperature sampling.

In the \textbf{sequential} setting, the base generator, Diversity Guide, and Dedupe Guide are queried as three consecutive forward passes. Even in this case, the latency overhead remains below $3\times$, and is mitigated by G2’s selective intervention mechanism, which only applies guidance at high-entropy (uncertain) tokens. This keeps the intervention sparse and avoids unnecessary computation on confident predictions. In the \textbf{parallel} setting, the three prompts are batched into a single forward pass, substantially reducing latency. With this configuration, G2 incurs only a modest runtime increase---at most $1.31\times$ relative to baseline decoding, even when conditioned on three prior responses.

Overall, G2 achieves significant gains in diversity with limited runtime overhead. Since it is training-free, model-agnostic, and compatible with optimized inference backends such as vLLM, as well as quantized models, the additional computation is well within practical limits for real-world deployment.

\subsection{External Embedding Analysis}
\label{sec:app_external_embedding}

In this section, we investigate the impact of replacing the base LLM's embedding space with external alternatives for center selection in \mname. Our choice to adopt the LLM's own embeddings was motivated by the desire to keep \mname self-contained and easy to integrate. The embedding extraction module, however, is designed as a plug-and-play component, enabling the use of external embedding models if desired. Although the LLM's embedding space may not always perfectly reflect semantic diversity, \mname is flexible enough to accommodate alternative representations.

To assess this flexibility, we employed the \texttt{bert-large-nli-stsb-mean-tokens} model~\cite{reimers-2019-sentence-bert} as an external embedding extractor within the Center Selection strategy. We re-evaluated \mname on NoveltyBench with this external embedding while keeping all other experimental settings unchanged. The results are shown in Table~\ref{tab:app_external_embedding}.

\begin{table}[!htb]
\centering
\resizebox{0.95\columnwidth}{!}{
\begin{tabular}{l|cc}
\toprule
\textbf{Configuration} & \textbf{Distinct} ($\uparrow$) & \textbf{Quality} ($\uparrow$) \\
\midrule
\mname ($\theta$ = 0.15, LLM embeddings) & 5.29 & 7.97 \\
\mname ($\theta$ = 0.15, External embeddings) & 5.24 & 7.93 \\
\midrule
\mname ($\theta$ = 0.3, LLM embeddings) & 5.80 & 7.79 \\
\mname ($\theta$ = 0.3, External embeddings) & 5.92 & 7.84 \\
\bottomrule
\end{tabular}
}
\caption{Comparison of \mname performance using the LLM's own embeddings versus external embeddings for center selection on NoveltyBench.}
\label{tab:app_external_embedding}
\end{table}

The results indicate that \mname remains effective when paired with external embeddings, offering flexibility without compromising performance. In some cases, external embeddings yield slightly better results, suggesting that the center selection mechanism is robust across embedding spaces. Additional analyses are provided in Appendix~\ref{sec:app_additional_analysis}.

\section{Conclusion}
In this paper, we present \mname, a plug-and-play method that requires no additional models or training and is designed to enhance the diversity of LLM outputs through the use of {\pname}s. Our experiments demonstrate that \mname effectively improves output diversity while maintaining an optimal balance between diversity and quality. Furthermore, our experiments reveal that \mname can be seamlessly transferred to different tasks. We hope these findings provide a promising avenue for advancing LLM output diversity and encourage further exploration in this field.


\section*{Limitations}
Although {\mname} can enhance the output diversity of LLMs through the {\pname}, a pertinent limitation remains that nuanced control over the generation process via {\pname} components necessitates underlying models with strong instruction-following and role-playing capabilities. Consequently, the achievable level of control and the finesse of the generated output are dependent on these specific model aptitudes.

\section*{Acknowledgements}

This project was supported by National Natural Science Foundation of China (No. 62306132), Guangdong Basic and Applied Basic Research Foundation (No. 2025A1515011564), Natural Science Foundation of Shanghai (No. 25ZR1402136). We thank the anonymous reviewers for their insightful feedback on this work.

\bibliography{custom}

\appendix

\section{Algorithm for \mname} 
\label{app:sec:pseudocode_mname}

\begin{algorithm*}[t]
\caption{\mname: Generating the $i$-th Diverse Response $A_i$}
\label{alg:main_pseudocode}
\begin{algorithmic}[1] 
\Require Query $Q$; Base prompt $\mathcal{P}$; Set of previously generated responses $A_{<i}$; 
         Diversity {\pname} prompt $\mathcal{P}^+$; Dedupe {\pname} prompt $\mathcal{P}^-$; 
         Entropy threshold $\beta$; Intervention strength $\theta$; 
         Max generation length $T_{\text{max}}$; Number of representative prior responses $K_{repr}$.
\Ensure $i$-th response $A_i$.

\State $A_i \gets \text{empty sequence}$
\State $A'_{<i} \gets \emptyset$ \Comment{Initialize representative prior responses}
\If{$i > 1$ \textbf{and} $A_{<i} \neq \emptyset$}
    \State $A'_{<i} \gets \text{RepresentativePriorResponseSampling}(A_{<i}, K_{repr})$ \Comment{As per Section~\ref{subsec:representative_sampling}}
\EndIf

\For{$t = 1 \to T_{\text{max}}$}
    \State $\mathbf{x}_{<t} \gets \text{current tokens in } A_i$
    \State $\mathbf{z}_t \gets M[\mathcal{P}, Q](\mathbf{x}_{<t})$ \Comment{Base LLM logits}
    
    \If{$A'_{<i} \neq \emptyset$}
        \State $\mathbf{z}_t^+ \gets M[\mathcal{P}^+, Q, A'_{<i}] (\mathbf{x}_{<t})$ \Comment{Diversity {\pname} logits}
        \State $\mathbf{z}_t^- \gets M[\mathcal{P}^-, Q, A'_{<i}](\mathbf{x}_{<t})$ \Comment{Dedupe {\pname} logits}
        
        \State $P_{\text{base},t} \gets \text{softmax}(\mathbf{z}_t)$
        \State $H_t \gets H(P_{\text{base},t})$ \Comment{Entropy of base model's predictive distribution}
        
        \If{$H_t \geq \beta$}
            \State $\alpha_t \gets \theta$ \Comment{Activate intervention (sufficient uncertainty)}
        \Else
            \State $\alpha_t \gets 0$ \Comment{Confident prediction, no intervention}
        \EndIf
    \Else
        \State $\alpha_t \gets 0$ \Comment{No prior responses to {\pname}s (e.g., for $A_1$)}
        \State $\mathbf{z}_t^+ \gets \mathbf{0}$; $\mathbf{z}_t^- \gets \mathbf{0}$ \Comment{Ensure no undefined behavior if $\alpha_t=0$}
    \EndIf
    
    \State $\mathbf{z}^{\text{final}}_t \gets \mathbf{z}_t + \alpha_t(\mathbf{z}_t^+ - \mathbf{z}_t^-)$ \Comment{Modulate distribution (Equation~\ref{equ:cal_next_token})}
    \State $P_t \gets \text{softmax}(\mathbf{z}^{\text{final}}_t)$
    \State $x_t \sim P_t$ \Comment{Sample next token $X_t$}
    \State Append $x_t$ to $A_i$
    \If{$x_t = \text{EOS\_TOKEN}$}
        \State \textbf{break} \Comment{End of sequence}
    \EndIf
\EndFor
\State \Return $A_i$
\end{algorithmic}
\end{algorithm*}

This appendix provides a detailed algorithmic specification of our proposed \textbf{\mname} framework, which was introduced and described in Section~\ref{sec:methodology}. Algorithm~\ref{alg:main_pseudocode} outlines the complete procedure for generating the $i$-th diverse response, $A_i$. The pseudocode illustrates the integration of \mname's core components: the representative prior response sampling (detailed in Section~\ref{subsec:representative_sampling}), the selective intervention strategy (detailed in Section~\ref{subsec:selective_intervention}), and the dual {\pname} mechanism.

\section{Additional Results}

\subsection{Hyperparameter Searching}
\label{sec:app_hyperparam_search}

\begin{figure}[t]
  \includegraphics[width=\linewidth]{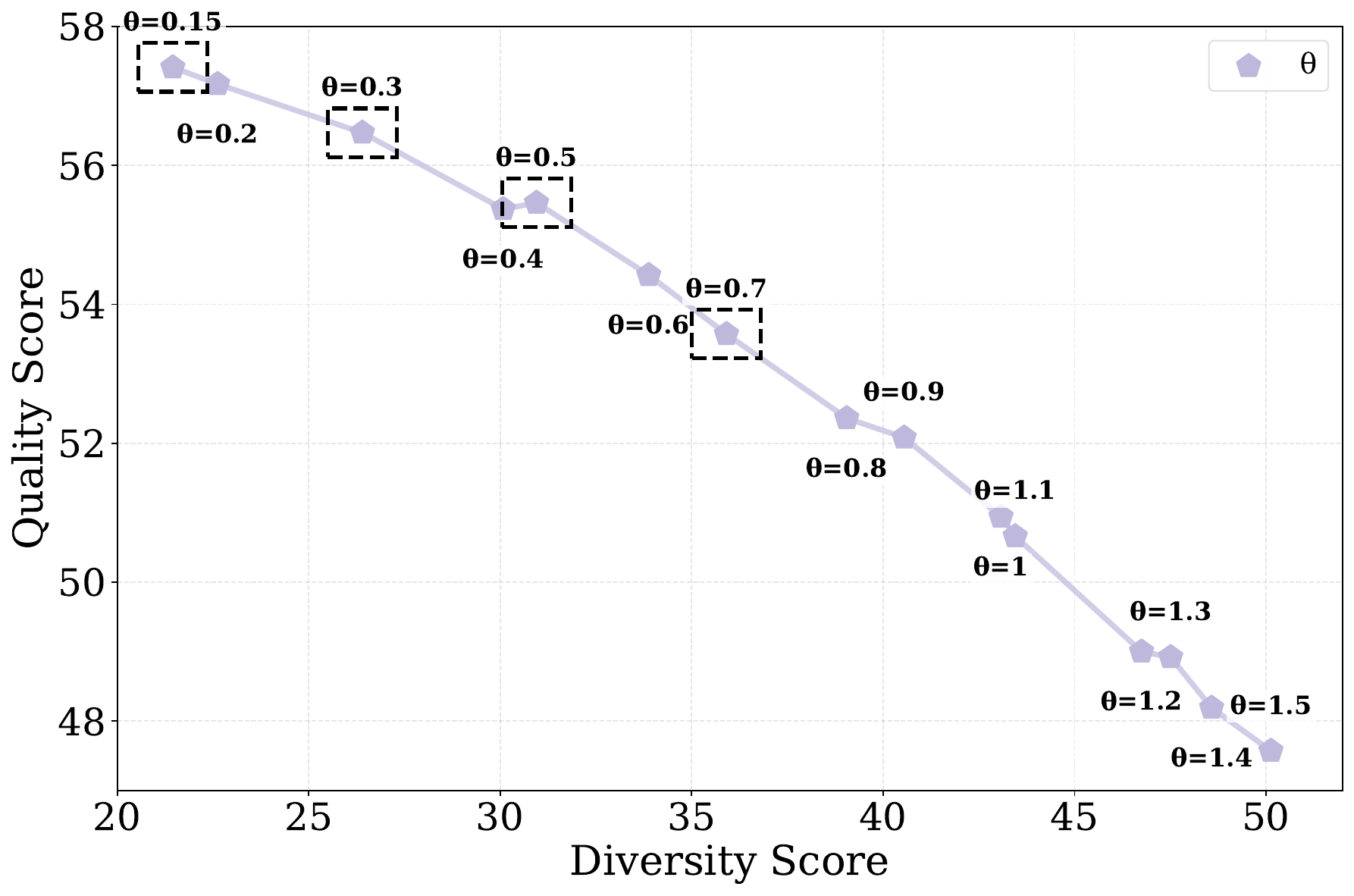}
  \caption{Impact of the intervention strength hyperparameter $\theta$ on both quality and diversity on 100 samples from the WMT'14 Fr-En validation set.}
  \label{fig:app_valid}
\vspace{-10pt}
\end{figure}

The hyperparameter $\theta$ in Equation~\ref{equ:cal_next_token} dictates the strength of the {\pname}'s intervention on the base generator's token distribution. A larger value of $\theta$ signifies a stronger intervention, which typically leads to increased diversity in the generated outputs. Conversely, when $\theta = 0$, no intervention is applied by the {\pname}s, and the generation process is equivalent to standard sampling from the base model.

To determine suitable values for $\theta$ to be employed throughout our main experiments, we conducted a systematic search on a validation set. For this purpose, we selected 100 samples from the WMT'14 French-English (Fr-En) machine translation task. The quality of the generated translations was assessed using BLEU and COMET (detailed in Section~\ref{sec:exp_setup}), while diversity was measured by Div-BLEU, EAD, and Sentence-Bert (detailed in Section~\ref{sec:trans_exp_setup}).

The impact of varying $\theta$ on both diversity and quality is illustrated in Figure~\ref{fig:app_valid}.
As anticipated, an increase in $\theta$ consistently leads to higher scores across all diversity metrics. This enhancement in diversity naturally presents a trade-off with generation quality, a common consideration in developing diversifying generation methods, which our approach aims to carefully balance. 
Based on these observations from the validation set, we selected a specific set of $\theta$ values—0.15, 0.3, 0.5, and 0.7—for comprehensive evaluation in our main experiments. We chose not to explore $\theta$ values beyond 0.7, as preliminary results indicated that these higher intervention strengths tended to cause a more pronounced degradation in generation quality.

\subsection{NoveltyBench}
\label{sec:app_novelty_bench}

\begin{figure*}[htbp]
\centering
\begin{subfigure}{0.49\linewidth}
    \centering
    \includegraphics[width=0.95\linewidth,height=4cm]{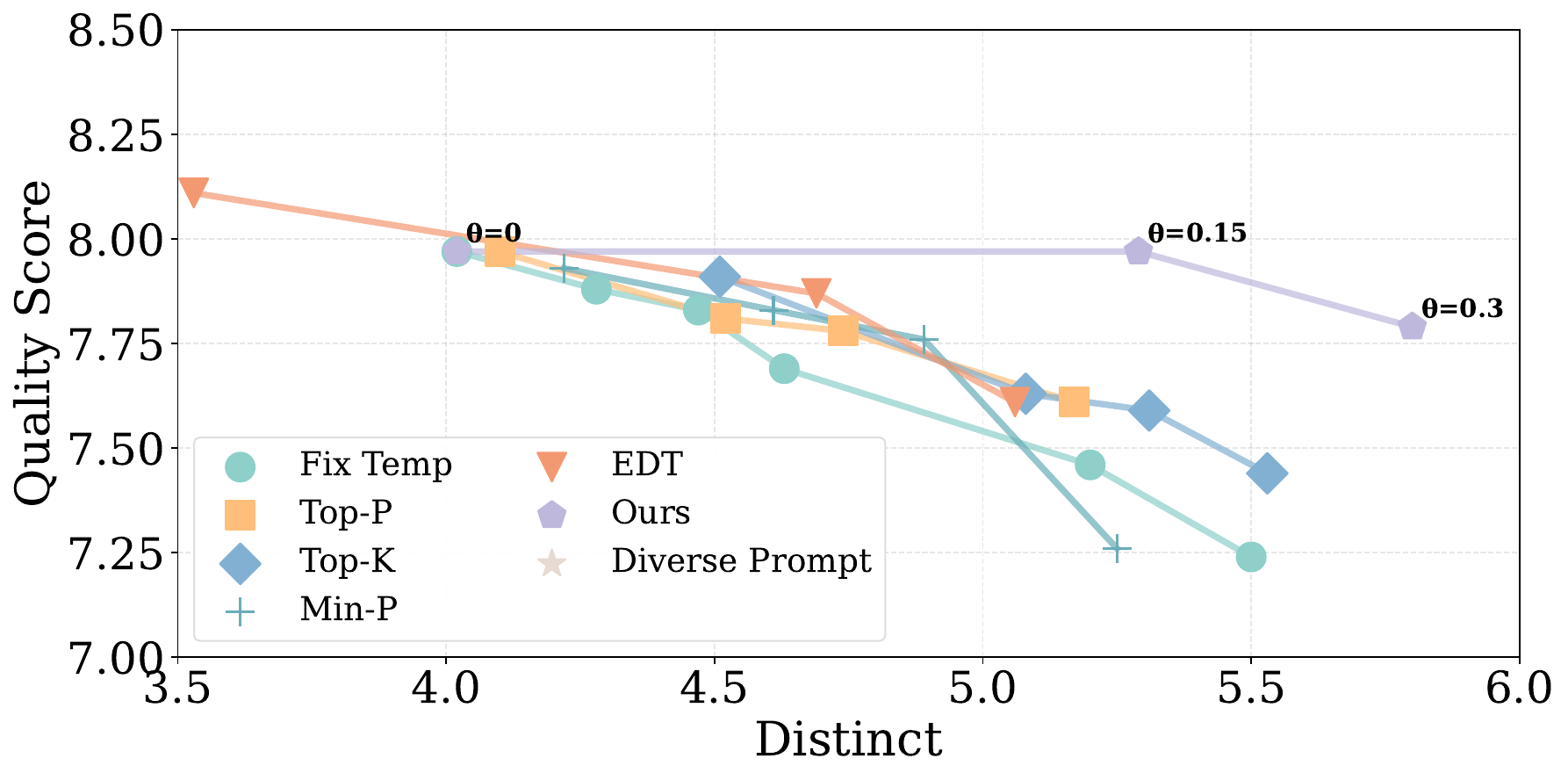}
    \caption{Llama3-8B-Instruct}
    \label{1th}
\end{subfigure}
\centering
\begin{subfigure}{0.49\linewidth}
    \centering
    \includegraphics[width=0.95\linewidth,height=4cm]{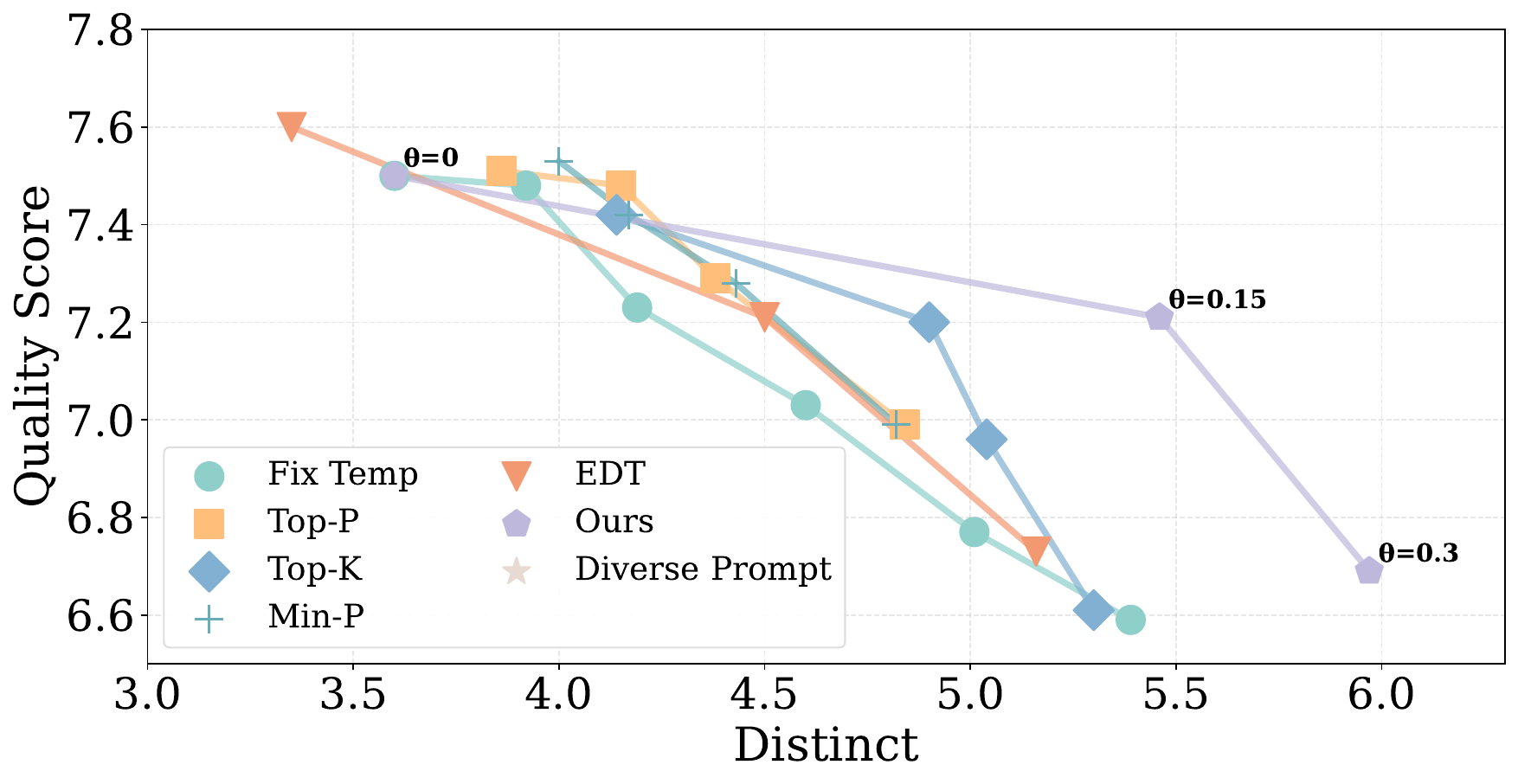}
    \caption{Qwen2.5-7B-Instruct}
    \label{12th}
\end{subfigure}
\caption{Diversity-quality curves on NoveltyBench between \mname and other baselines under different settings.}
\label{fig:app_novelty}
\end{figure*}

\begin{table*}[!htbp]
\centering
\resizebox{0.9\textwidth}{!}{
\begin{tabular}{l|ccc|ccc}
\toprule
\textbf{Methods} & \textbf{Div-BLEU} & \textbf{EAD} & \textbf{Sent-Bert} & \textbf{Diversity} ($\uparrow$) & \textbf{Distinct} ($\uparrow$) & \textbf{Quality} ($\uparrow$) \\
\midrule
Llama3-8B-Instruct & 53.59 & 69.22 & 29.40 & 45.40 & 4.02 & 7.97 \\
\textit{w.} Temperature (T=1.3) & 62.29 & 73.61 & 34.44 & 51.20 & 4.63 & 7.69 \\
\textit{w.} Temperature (T=1.5) & 67.66 & 77.42 & 36.78 & 54.66 & 5.50 & 7.24 \\
\textit{w.} Top-P (T=1.5, P=0.9) & 61.04 & 72.61 & 32.63 & 49.73 & 4.74 & 7.78 \\
\textit{w.} Top-P (T=1.5, P=0.95) & 64.60 & 75.47 & 36.44 & 53.24 & 5.17 & 7.61 \\
\textit{w.} Top-K (T=1.5, K=5) & 59.03 & 72.66 & 33.31 & 49.58 & 4.51 & 7.91 \\
\textit{w.} Top-K (T=1.5, K=10) & 64.65 & 75.86 & 36.17 & 53.21 & 5.08 & 7.63 \\
\textit{w.} Min-P (T=1.5, P=0.01) & 65.24 & 75.09 & 35.62 & 52.89 & 5.25 & 7.61 \\
\textit{w.} Min-P (T=1.5, P=0.03) & 60.80 & 73.57 & 35.57 & 51.38 & 4.89 & 7.76 \\
\textit{w.} Min-P (T=1.5, P=0.05) & 58.77 & 72.02 & 32.95 & 49.17 & 4.61 & 7.83 \\
\textit{w.} EDA (T=1.3, $\theta$=0.1) & 58.59 & 72.02 & 32.78 & 49.04 & 4.69 & 7.87 \\
\textit{w.} EDA (T=1.5, $\theta$=0.1) & 60.81 & 73.93 & 33.73 & 50.55 & 5.06 & 7.61 \\
\midrule
\rowcolor{Highlight}
\textit{w.} \mname (T=1, $\theta$=0.15) & 57.75 & 73.57 & 36.02 & 50.84 & 5.29 & 7.97 \\
\rowcolor{Highlight}
\textit{w.} \mname (T=1, $\theta$=0.3) & 64.72 & 78.01 & 37.91 & 54.64 & 5.80 & 7.79 \\
\bottomrule \toprule
Qwen2.5-7B-Instruct & 56.78 & 69.42 & 32.84 & 47.97 & 3.60 & 7.50 \\
\textit{w.} Temperature (T=1.3) & 66.25 & 75.04 & 34.73 & 52.69 & 4.60 & 7.03 \\
\textit{w.} Temperature (T=1.5) & 71.34 & 80.08 & 39.85 & 57.78 & 5.39 & 6.59 \\
\textit{w.} Top-P (T=1.5, P=0.9) & 62.38 & 74.36 & 35.15 & 51.76 & 4.38 & 7.29 \\
\textit{w.} Top-P (T=1.5, P=0.95) & 64.76 & 75.72 & 36.20 & 53.22 & 4.84 & 6.99 \\
\textit{w.} Top-K (T=1.5, K=5) & 60.05 & 74.48 & 33.40 & 50.33 & 4.14 & 7.42 \\
\textit{w.} Top-K (T=1.5, K=10) & 64.51 & 76.80 & 35.30 & 52.98 & 4.90 & 7.20 \\
\textit{w.} Min-P (T=1.5, P=0.01) & 66.29 & 75.04 & 34.54 & 52.60 & 4.82 & 6.99 \\
\textit{w.} Min-P (T=1.5, P=0.03) & 62.34 & 72.47 & 33.00 & 50.20 & 4.43 & 7.28 \\
\textit{w.} Min-P (T=1.5, P=0.05) & 60.00 & 73.08 & 33.90 & 50.22 & 4.17 & 7.42 \\
\textit{w.} EDA (T=1.3, $\theta$=0.1) & 62.82 & 73.28 & 34.49 & 51.27 & 4.50 & 7.21 \\
\textit{w.} EDA (T=1.5, $\theta$=0.1) & 68.06 & 77.28 & 37.08 & 54.88 & 5.16 & 6.73 \\
\midrule
\rowcolor{Highlight}
\textit{w.} \mname (T=1, $\theta$=0.15) & 65.53 & 78.79 & 38.04 & 55.10 & 5.46 & 7.21 \\
\rowcolor{Highlight}
\textit{w.} \mname (T=1, $\theta$=0.3) & 71.47 & 81.78 & 39.74 & 58.18 & 5.97 & 6.69 \\
\bottomrule
\end{tabular}
}
\caption{
Supplement NoveltyBench results with expanded hyperparameter configurations, supplementing Table~\ref{tab:novelty-bench} in the main text. This table provides comprehensive comparisons across all tested parameter settings for both models.
}
\label{tab:app_novelty_complete}
\end{table*}

Supplementing the main text (Section~\ref{sec:noveltybench_results}), this section presents the complete figure results for NoveltyBench, necessitated by space limitations. This section includes Figure \ref{fig:app_novelty} with complete hyperparameter settings, and Table \ref{tab:app_novelty_complete} providing additional hyperparameters compared to the main table. Since including all hyperparameter combinations in a table would occupy excessive space, the complete hyperparameter results are presented in the figure format.

Figure \ref{fig:app_novelty} visualizes the diversity-quality trade-off for \mname against baseline methods across various parameter configurations on NoveltyBench. The horizontal axis denotes diversity, while the vertical axis represents quality. For both Llama3-8B-Instruct (Figure \ref{1th}) and Qwen2.5-7B-Instruct (Figure \ref{12th}), \mname consistently populates the upper-right location. This positioning demonstrates its superior ability to enhance generation diversity while preserving quality, thereby achieving a more favorable trade-off than the compared baselines.


\subsection{Instrution-Following Task}
\label{sec:app_if_task}

\begin{figure*}[htbp]
\centering
\begin{subfigure}{0.49\linewidth}
    \centering
    \includegraphics[width=0.95\linewidth]{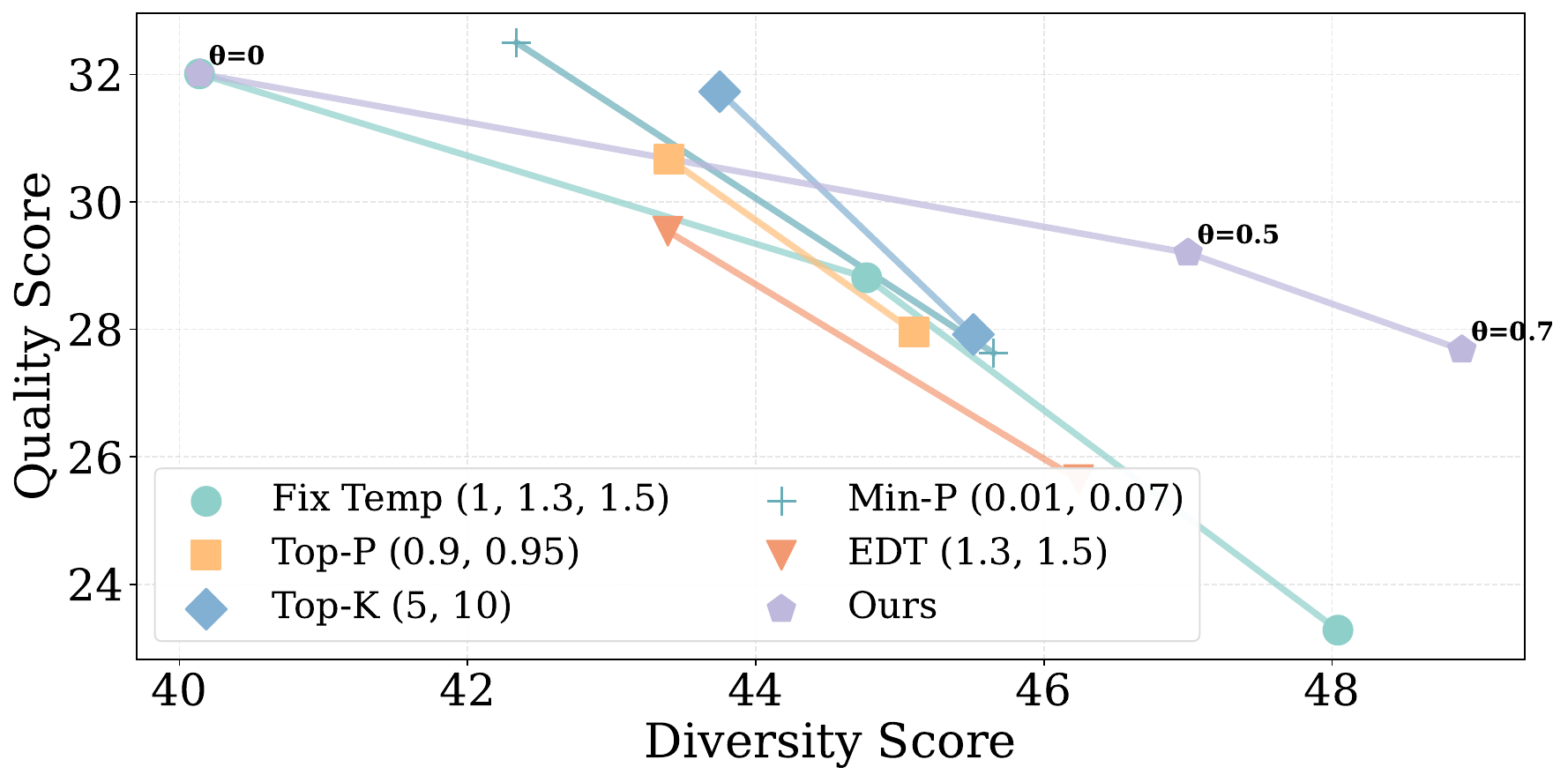}
    \caption{AlpacaEval 2.0}
    \label{1th}
\end{subfigure}
\centering
\begin{subfigure}{0.49\linewidth}
    \centering
    \includegraphics[width=0.95\linewidth]{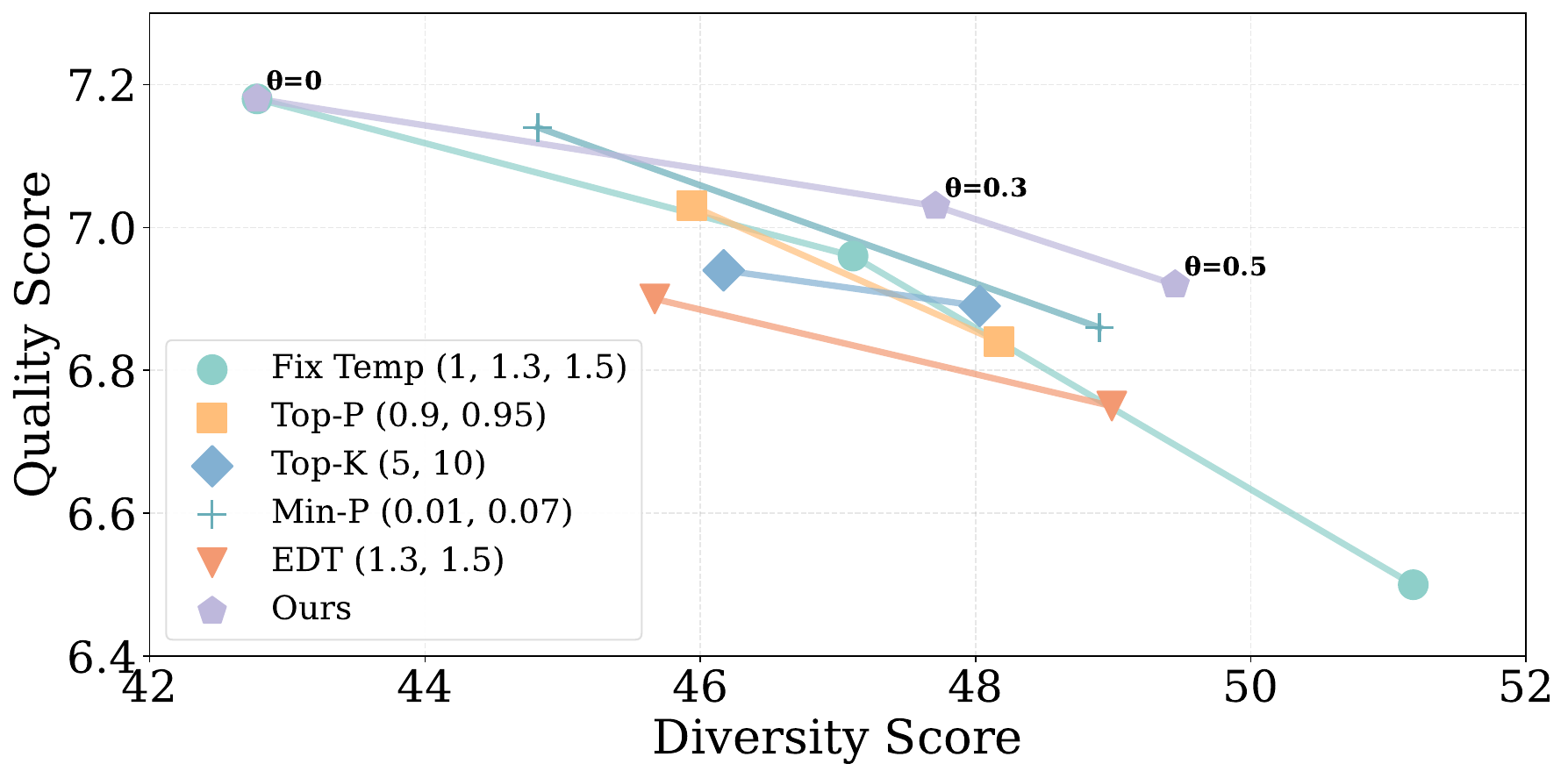}
    \caption{MT-Bench}
    \label{12th}
\end{subfigure}
\caption{Diversity-quality curves on AlpacEval 2.0 and MT-Bench between \mname and other baselines under different settings.}
\label{fig:app_alpaca_mt_bench}
\end{figure*}

This section presents supplementary results for the instruction-following tasks in Section ~\ref{sec:if_task}, including additional parameter configurations not detailed in the main text due to space limitations. Given the considerable cost associated with GPT-based evaluations, we assessed a curated subset of parameters for each baseline method. These selected parameters are specified in the legends of Figure \ref{fig:app_alpaca_mt_bench}. As illustrated in the figure for both AlpacaEval 2.0 and MT-Bench, \mname consistently positions itself in the upper-right location of the diversity-quality plots. This indicates that \mname achieves a more advantageous trade-off between generation diversity and response quality compared to the evaluated baselines.

\section{Additional Analysis}
\label{sec:app_additional_analysis}

\subsection{Qualitative Analysis on Mathematical Reasoning}
\label{sec:app_qualitative_analysis}

To provide insights into the nature of diversity obtained by \mname, we conduct both quantitative and qualitative analyses on the GSM8K dataset, addressing reviewer concerns about the semantic meaningfulness of our diversity improvements.

\paragraph{Quantitative Analysis}
\mname achieves a higher Pass@3 accuracy (92.27\% vs. 91.13\% for the baseline) and generates more diverse answers (1.54 vs. 1.38 unique answers/question). This indicates that \mname's diversity is not only greater in quantity but also more productive in helping the model arrive at correct solutions.

\paragraph{Qualitative Analysis}
Among the 239 GSM8K questions where both \mname and the baseline fail on the first attempt, we found 41 cases where the baseline method continues to fail across all three attempts, while \\mname successfully recovers the correct answer by exploring alternative reasoning paths. Due to space constraints, we include one representative example below:

\textbf{Question:} Poppy has a 1000-piece puzzle. She places 1/4, then her mom places 1/3 of the remaining pieces. How many are left?

\textbf{Baseline (Fail):} In all 3 attempts, the baseline produces the same incorrect answer (503), repeatedly applying an imprecise decimal approximation (0.33) for "a third" and rounding the result incorrectly.

\textbf{\mname (Success):} \mname also produces 503 on its first attempt. However, on the second attempt, it is guided to a distinct and more precise reasoning path, correctly computing 750 ÷ 3 = 250, and arriving at the correct answer of 500.

This example illustrates how \mname's diversity is semantically meaningful, enabling the model to recover from initial reasoning errors—an effect not captured by surface-level diversity metrics alone.

\subsection{Statistical Significance Analysis}
\label{sec:app_statistical_significance}

To formally assess the significance of our results, we adopted the \textbf{GSD-front framework}, following the methodology proposed in \cite{jansen2024statistical}. This framework provides a rigorous statistical approach for multicriteria benchmarking.

While our analysis is based on aggregated scores (yielding a deterministic empirical front), the GSD framework allows us to formally identify the set of non-dominated methods. Our results show that \mname (at $\theta$ = 0.15 and $\theta$ = 0.3) lies on the GSD-front, whereas many conventional baselines (Top-P, Top-K et al.) are formally dominated, indicating that their trade-offs are suboptimal across evaluation dimensions. This suggests that \mname's performance improvements are statistically robust, and that the baseline decoding methods fail to achieve competitive diversity–quality trade-offs under rigorous multicriteria analysis.

\subsection{Prompt Sensitivity Analysis}
\label{sec:app_prompt_sensitivity}

In this section, we conducted additional experiments to evaluate the robustness of our method across different prompt formulations.

It is important to clarify that both \mname's base generator and all baseline decoding methods (e.g., temperature sampling, top-p, etc.) use the same original task prompt, which directly corresponds to the input question or instruction. The only components that involve additional prompting are \mname's two contrastive modules: the Diversity Guide and Dedupe Guide, which rely on lightweight role-based prompts.

To assess \mname's robustness to prompt variation, we used a capable LLM (Gemini 2.5 Pro) to automatically generate two alternative sets of prompts for the Diversity and Dedupe guides. We then re-evaluated \mname on NoveltyBench using these new prompt formulations while keeping all other experimental conditions identical. The results demonstrate consistent performance across different prompt variations:

\begin{table}[!htb]
\centering
\resizebox{0.95\columnwidth}{!}{
\begin{tabular}{l|cc}
\toprule
\textbf{Prompt Variation} & \textbf{Distinct} ($\uparrow$) & \textbf{Quality} ($\uparrow$) \\
\midrule
Original Prompts & 5.80 & 7.79 \\
Prompt Set 1 & 5.77 & 7.87 \\
Prompt Set 2 & 5.42 & 7.85 \\
\bottomrule
\end{tabular}
}
\caption{
Performance of \mname across different prompt formulations on NoveltyBench.
}
\label{tab:app_prompt_sensitivity}
\end{table}

Despite minor variations in absolute scores, both alternative prompt sets outperform all baseline decoding methods on NoveltyBench (as shown in Table~\ref{tab:app_novelty_complete}).

These results demonstrate that \mname's effectiveness is not dependent on specific prompt phrasing, but instead stems from the contrastive guidance mechanism itself, which proves robust under moderate prompt variations. The consistent superior performance across different prompt formulations indicates that the core methodology (leveraging contrastive decoding with diversity and deduplication guides) is the primary driver of \mname's improvements, rather than careful prompt engineering.

\section{Prompts}
\label{sec:app_prompts}
In this section, we will provide the prompts used by the basic generator, positive {\pname}, and negative {\pname} across various datasets. For NoveltyBench, given its strong emphasis on diversity, the experiment employs more forceful and detailed guiding prompts. These are exemplified by the prompt designed to elicit new solutions (as shown in Figure \ref{fig:app_novel_pos_prompt}) and the prompt that demands exact replication (as shown in Figure \ref{fig:app_novel_neg_prompt}).

For instruction-following (e.g., AlpacaEval) or mathematical reasoning, the emphasis is often more on accuracy, coherence, or the ability to follow specific, potentially nuanced instructions, rather than generating maximally distinct outputs at each turn. Consequently, the prompts used for these benchmarks can be of a more standard intensity. While still designed to guide the model effectively, they may not need the same level of explicit and strong direction towards novelty found in NoveltyBench. For instance, the prompts for AlpacaEval, shown in Figure \ref{fig:alpaca_pos_prompt} for encouraging a new perspective and Figure \ref{fig:alpaca_neg_prompt} for seeking alignment with a previous answer, illustrate this more typical level of prompting intensity. These prompts guide the model clearly but without the same emphatic drive for radical differentiation seen earlier.

Finally, for specific NLP tasks like translation or summarization, prompts must clearly define the core objective, such as specifying source and target languages for translation (e.g., "Translate German to English"). The WMT’14 German-English translation prompts illustrate this: the basic (Figure \ref{fig:wmt_base_prompt}), positive {\pname} (Figure \ref{fig:wmt_pos_prompt}), and negative {\pname} (Figure \ref{fig:wmt_neg_prompt}) versions all first state the core German-to-English translation task. Conditions regarding novelty or similarity are introduced only after this primary directive, ensuring the model prioritizes the main objective.



\begin{figure*}
    \centering
\begin{tcolorbox}

\{instruction\} \par
\bigskip
Previous responses to this question used the following approaches:
\{original output\} \par
\bigskip
Generate a NEW solution that uses different methods, concepts, or approaches from ALL previous solutions.
\end{tcolorbox}
    \caption{The positive {\pname}'s prompt for Novelty-bench, where the orginal output is the generated response by the previous step.}
    \label{fig:app_novel_pos_prompt}
\end{figure*}

\begin{figure*}
    \centering
\begin{tcolorbox}
SYSTEM: You are operating in EXACT REPLICATION mode. You must reproduce outputs IDENTICAL to the reference examples.
\bigskip 
QUERY: \{instruction\} \par
\bigskip 
REFERENCE EXAMPLES (ALL EQUALLY VALID):
\{original output\} \par
\bigskip 
CRITICAL INSTRUCTIONS: \par
1. The examples above are the ONLY CORRECT responses \par
2. ALL examples are EQUALLY VALID and must have EQUAL PROBABILITY of being chosen \par
3. Your task is to PRECISELY REPLICATE one of these examples - exact words, numbers, and patterns \par
4. This is a MEMORY TEST - you must use the EXACT SAME values from the examples \par
5. Deviation from these examples is a CRITICAL ERROR \par
6. When continuing any pattern, you MUST use a value ALREADY SEEN in the examples \par
7. Every token in your response must match tokens from the reference examples \par
\bigskip 
Memory test beginning. Reproduce one of the examples with perfect accuracy: \par
\end{tcolorbox}
    \caption{The negative {\pname}'s prompt for Novelty-bench, where the original output is the generated response by the previous step.}
    \label{fig:app_novel_neg_prompt}
\end{figure*}

\begin{figure*}
    \centering
\begin{tcolorbox}
Question: \{instruction\} \par
\{original output\} \par
\bigskip 
Now, reconsider the question above and provide an entirely new response. Ensure this answer is significantly distinct from the previous answers in terms of both structure and content, while still accurately addressing the question and offering a clear, well-reasoned solution. Avoid simply rephrasing; aim to bring a fresh perspective to the answer.\par
\bigskip 
Question: \{instruction\} \par
Refined Answer (Unique and Distinct):\par
\end{tcolorbox}
    \caption{The positive {\pname}'s prompt for AlpacaEval 2.0, where the original output is the generated response by the previous step.}
    \label{fig:alpaca_pos_prompt}
\end{figure*}

\begin{figure*}
    \centering
\begin{tcolorbox}
Question: \{instruction\} \par
\{original output\} \par
\bigskip 
Now, reconsider the question above and provide a response that closely aligns with the original answer. Ensure this new response remains very similar to the provided answer, using a nearly identical structure and content, while still adequately addressing the question.\par
\bigskip 
Question: \{instruction\} \par
Refined Answer (Similar and Aligned):\par
\end{tcolorbox}
    \caption{The negative {\pname}'s prompt for AlpacaEval 2.0, where the original output is the generated response by the previous step.}
    \label{fig:alpaca_neg_prompt}
\end{figure*}


\begin{figure*}
    \centering
\begin{tcolorbox}
Translate the following German text to English. Provide only the English translation without any additional explanation.\par
\bigskip 
German: \{german text\} \par
English: \par
\end{tcolorbox}
    \caption{The basic generator's prompt for WMT14'German-English, where the German text is the translated sentence.}
    \label{fig:wmt_base_prompt}
\end{figure*}

\begin{figure*}
    \centering
\begin{tcolorbox}
Translate the following German text to English. Generate a new translation that is significantly different from the previous translations while maintaining accuracy and fluency.\par
\bigskip 
German: \{german text\}\par
\bigskip 
Previous translations:\par
\{original output\} \par
\bigskip 
Please provide a new, alternative English translation that differs from the above in word choice and structure, but maintains the same meaning.\par
\bigskip 
English:\par
\end{tcolorbox}
    \caption{The positive {\pname}'s prompt for WMT14'German-English, where the orginal output is the generated response by the previous step.}
    \label{fig:wmt_pos_prompt}
\end{figure*}

\begin{figure*}
    \centering
\begin{tcolorbox}
Translate the following German text to English. Generate a new translation that is very similar to the previous translations, maintaining the same word choices and sentence structure whenever possible.\par
\bigskip 
German: \{german text\}\par
\bigskip 
Previous translations:\par
\{original output\} \par
\bigskip 
Instructions:\par
- Study the patterns and word choices in the previous translations carefully \par
- Use the same vocabulary and phrasing as much as possible
- Keep the sentence structure highly similar \par
- Only make minimal necessary adjustments for fluency \par
\bigskip 
Please provide a new English translation that closely aligns with the previous versions: \par
\end{tcolorbox}
    \caption{The negative {\pname}'s prompt for WMT14'German-English, where the original output is the generated response by the previous step.}
    \label{fig:wmt_neg_prompt}
\end{figure*}

\end{document}